\documentclass{ieeeaccess}
\usepackage{cite}
\usepackage{amsmath,amssymb,amsfonts}
\usepackage{algorithmic}
\usepackage{graphicx}
\usepackage{textcomp}
\usepackage{booktabs}
\usepackage{multirow}
\usepackage{amsmath}
\usepackage{mathtools}
\usepackage{color}
\usepackage{ulem}
\def\BibTeX{{\rm B\kern-.05em{\sc i\kern-.025em b}\kern-.08emT\kern-.1667em\lower.7ex\hbox{E}\kern-.125emX}}
\begin{document}
\doi{10.1109/ACCESS.2017.DOI}

\title{Exploitation of Channel-Learning for Enhancing 5G Blind Beam Index Detection}
\author{\uppercase{Ji Yoon~Han}\authorrefmark{1}, \IEEEmembership{Student Member, IEEE},
\uppercase{Ohyun Jo}\authorrefmark{2}, \IEEEmembership{Member, IEEE},
and \uppercase{Juyeop Kim}\authorrefmark{3}\IEEEmembership{Member, IEEE}.}
\address[1]{Department of Electronics Engineering, Sookmyung Women's University, South Korea (e-mail: hanjiyoon17@sookmyung.ac.kr)}
\address[2]{Department of Computer Science, Chungbuk National University, South Korea (e-mail: ohyunjo@chungbuk.ac.kr)}
\address[3]{Department of Electronics Engineering, Sookmyung Women's University, South Korea (e-mail: jykim@sookmyung.ac.kr)}
\tfootnote{This research was supported by Sookmyung Women's University Research Grants (1-1803-2005), the Basic Science Research Program through the National Research Foundation of Korea (NRF) funded by the MSIP (2018R1C1B5045), and Institute for Information \& communications Technology Promotion (IITP) grant funded by the Korea government(MSIP) (2018-0-00726, Development of Software-Defined Cell/Beam Search Technology for Beyond-5G Systems). }

\markboth
{J. Y. Han \headeretal: Submission for IEEE Access}
{J. Y. Han \headeretal: Submission for IEEE Access}

\corresp{Corresponding author: Juyeop Kim (e-mail: jykim@sookmyung.ac.kr) and Ohyun Jo (e-mail: ohyunjo@chungbuk.ac.kr) .}

\begin{abstract}
Proliferation of 5G devices and services has driven the demand for wide-scale enhancements ranging from data rate, reliability, and compatibility to sustain the ever increasing growth of the telecommunication industry. In this regard, this work investigates how machine learning technology can improve the performance of 5G cell and beam index search in practice. The cell search is an essential function for a User Equipment (UE) to be initially associated with a base station, and is also important to further maintain the wireless connection. Unlike the former generation cellular systems, the 5G UE faces with an additional challenge to detect suitable beams as well as the cell identities in the cell search procedures. Herein, we propose and implement new channel-learning schemes to enhance the performance of 5G beam index detection. The salient point lies in the use of machine learning models and softwarization for practical implementations in a system level.  We develop the proposed channel-learning scheme including algorithmic procedures and corroborative system structure for efficient beam index detection. We also implement a real-time operating 5G testbed based on the off-the-shelf Software Defined Radio (SDR) platform and conduct intensive experiments with commercial 5G base stations. The experimental results indicate that the proposed channel-learning schemes outperform the conventional correlation-based scheme in real 5G channel environments.
\end{abstract}


\titlepgskip=-15pt

\maketitle

\section{Introduction}
\label{sec:introduction}
\PARstart{B}{eyond} the Long Term Evolution (LTE), the evolved 5G mobile communication has newly taped out its commercial service since 2019.
The 5G aims to provide enhanced service experience in the aspects of capacity, latency, reliability and massive connectivity.
The 5G system includes the groundbreaking technologies which differentiate the 5G from the LTE.
Representatively, the 5G enables multi-beam transmission, provides flexible frame structure and supports wider bandwidth in millimeter wave frequency bands \!\!\cite{b1}.
These new enabler technologies can be realized in commercial UEs and network devices to provide the innovative 5G applications.

However, even after the 5G commercial service has begun, it is still essential to make much effort for stabilizing the operation of each new technology in practice.
One of the new 5G technologies that we need to focus on is the cell search procedure.
The cell search procedure basically identifies the starting point of a symbol when the UE initially tries to connect to a base station.
After the cell search, the UE finds the symbol timing and the Physical Cell ID (PCI) of the base station, which are utilized for further decoding processes.
If the UE fails to search the correct cell and its symbol boundary, it may occur a critical impediment to the 5G system in a coverage aspect because the 5G system generally utilizes higher range of frequency bands and suffers more severe signal attenuation. The 5G system therefore requires highly sophisticated schemes for improving the performance of the cell search and extending its coverage without deploying additional base stations.

As the cell search is important in terms of communication between a base station and a UE, cell search issues are dealt throughout the generations.
The research normally aims to improve the detection probability of the cell search and to reduce the computational complexity.
A preamble sequence which is specially constructed for low-complex detection is proposed in \cite{a3}.
A novel algorithm for Primary Synchronization Signal (PSS) detection is devised in \cite{a1}, and the primary group identifier is suggested in \cite{a2} for LTE cell search procedure. The aforementioned works offer efficient ways for PSS detection, which is the most significant sub-process for ultimate success in the cell search.
In addition, the detailed target cell scenarios in 3G and LTE systems are surveyed in \cite{a8}.

One of the key issues in the 5G cell search is to find a proper beam between a base station and a UE \!\cite{qualcomm}.
According to the 5G specification \!\cite{TS38304}, the UE should find a suitable beam during the cell search by inspecting received synchronization signal.
Each 5G synchronization signal contains an index of the corresponding beam, and the UE can recognize the received beam by deriving its index from the synchronization signal.

Especially, millimetre Wave (mmWave) communications significantly depends on transmission direction so the cell search should be proceeded across wide range of angular space. The several research works are presented for the cell search procedure in mmWave bands. 
In \cite{a4}, new design options are considered for transmitting and receiving synchronization signals in a base station aspect.
The effect of various beamforming schemes between a base station and a UE are considered in \cite{a5}, which concludes that the use of digital beamforming for initial access will expedite the whole process significantly.
In \cite{a7} and \cite{a6}, the use of position information provided by UE is investigated for improving the performance of the cell search.
In detail, \cite{a6} uses a geo-located context database which stores the position information over time for provisioning future search procedures. 

The beam search issue becomes more challenging when it comes to initial cell selection, where the UE has no serving cell and cannot use any preliminary information.
Thus, the UE should detect the synchronization signal in a blind way.
This blind detection may work properly in high Signal to Noise Ratio (SNR) environments, but a sophisticated detection scheme should be applied in low SNR environments.
The effect of noise becomes dominant when SNR is low, so the received synchronization signal is likely to be critically distorted by the noise, which disturbs the blind detection process.

To solve the aforementioned challenge in 5G systems, Machine Learning (ML) techniques can be proper solutions which are emerging to deal with 5G issues. The ML techniques can infer and provide appropriate parameters by considering channel environments in an efficient way.
In \cite{a9}, the state-of-art ML-based techniques are discussed to deal with the issues of heterogeneous networks (HetNets), such as self-configuration, self-healing, and self-optimization.
\cite{a10} -- \cite{a12} also deal with emerging 5G challenges that can be overcome by the ML technologies.
In addition, there are a number of research works that propose applicable ML-based schemes in cellular systems.
For instance, an ML-based context-aware algorithm for initial access is proposed in a 5G mmWave system \cite{a13}.
\cite{a14} proposes a deep learning-based pilot assignment scheme for a massive multiple-input multiple-output system that supports multiple users. 
\!\cite{a15} proposes mapping functions and applies them to a deep learning model to enhance prediction of mmWave beam and blockage. 
Also, a deep learning solution was proposed in \cite{a16} for faster and more precise initial access in 5G mmWave networks.

In this paper, we describe all the technical aspects that compose the 5G cell search, and show the practical details including its implementation and experimental results. We first start with comprehensible explanations of the cell and beam index search procedure in 5G standards. We then investigate how the blind detection of a beam index can be improved by channel-learning schemes through mathematical formulations. Then we present an in-depth description of the practical implementation for the 5G cell and beam index search which is softwarized in full scale through the off-the-shelf Software Defined Radio (SDR) platforms. Lastly, we evaluate the performance at a system level through extensive experiments, and discuss to deduce insight and applicability of the channel-learning schemes.

\begin{figure}
\includegraphics[width=\columnwidth]{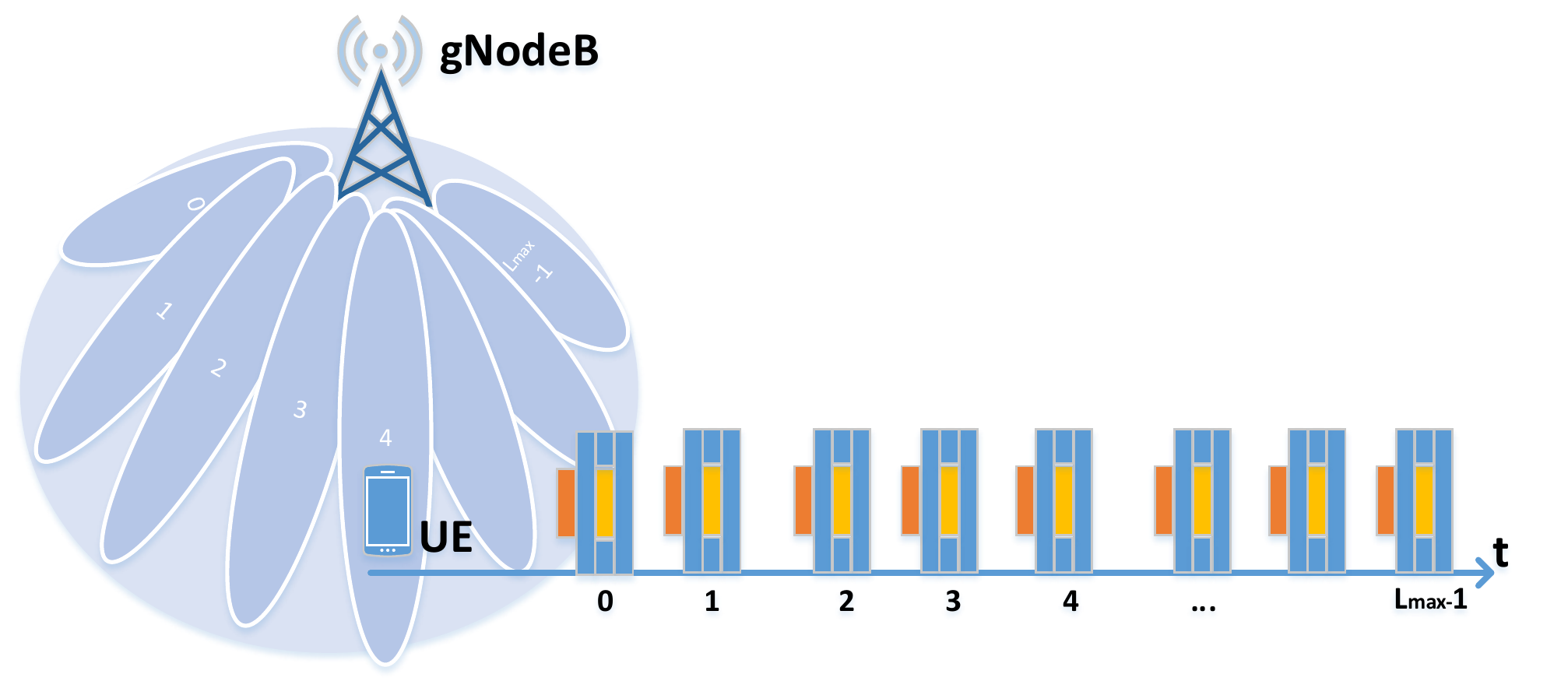}
\caption{The system model}%
\label{fig_1}
\end{figure}

\section{System Model Description}
Fig.\ref{fig_1} shows the overall system model that we consider in this paper. 
We assume that a 5G UE is in an initial cell selection scenario defined in \cite{TS38304}.
The UE does not have a serving cell and needs to quickly find a suitable cell to camp on. 
The UE does not also have prior knowledge about nearby cells and is required to search a cell blindly for whole Radio Frequency (RF) channels supported by the UE. 
In addition, we assume that there is a suitable 5G gNodeB (gNB) which is nearby the UE and whose signal is received by the UE with a sufficient power level.
The gNB forms its cell coverage with $L_{max}$ beams and each beam covers a certain area which is disjoint with the other beam coverages.
The UE tries the initial cell selection in the cell coverage, and consequently in one of the beam coverages, so it can receive only one kind of beam signal at the moment of the initial cell selection. 

\begin{figure}
\includegraphics[width=\columnwidth]{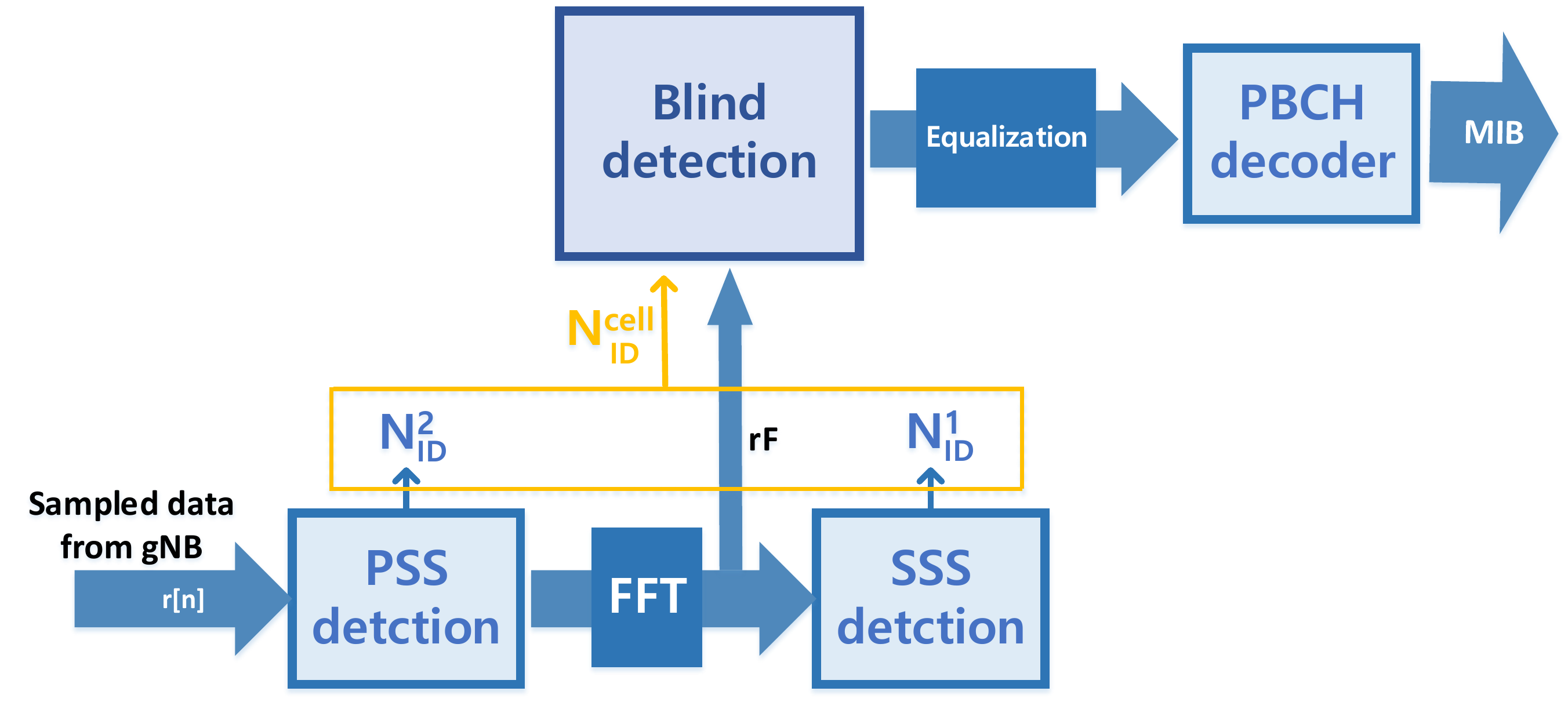}
\caption{The block diagram of the overall cell search process}%
\label{fig_2}
\end{figure}

At the beginning of the initial cell selection procedure, the UE selects an RF channel and proceeds {\it cell search} for inspecting if signal is received from nearby cells in the RF channel.
As shown in Fig.\ref{fig_2}, the cell search procedure is equivalent to synchronization which aims to identify the starting point of a symbol and basic cell configuration parameters by detecting a bundle of predetermined signals called a Synchronization Signal Block (SSB) \cite{SSB}.
An SSB includes two preambles; PSS and Secondary Synchronization Signal (SSS), whose sequences are defined in \cite{TS38211}.
There are 3 candidate sequences for the PSS and 336 candidate sequences for the SSS. 
The SSB also contains Physical Broadcast CHannel (PBCH) which carries the general system information called Master Information Block (MIB).

After detecting PSS, UE obtains the frame boundary of the gNB and the value of $N_{ID}^{2}$ from the index of the detected PSS. 
As a result of SSS detection, it gets the value of $N_{ID}^{1}$ from the index of the detected SSS. 
UE can consequently compute the PCI of the gNB, denoted by $N_{ID}^{cell}$, by the following equation,
\begin{equation}
N_{ID}^{cell} = N_{ID}^{2} + 3N_{ID}^{1}.
\end{equation}
Here, $N_{ID}^{1}$ ranges from \{0,1,...,335\} and $N_{ID}^{2}$ ranges from \{0,1,2\}. 
With the derived value of $N_{ID}^{cell}$, UE can further process to decode the PBCH and figure out the system information of the gNB.

\begin{figure}
\includegraphics[width=\columnwidth]{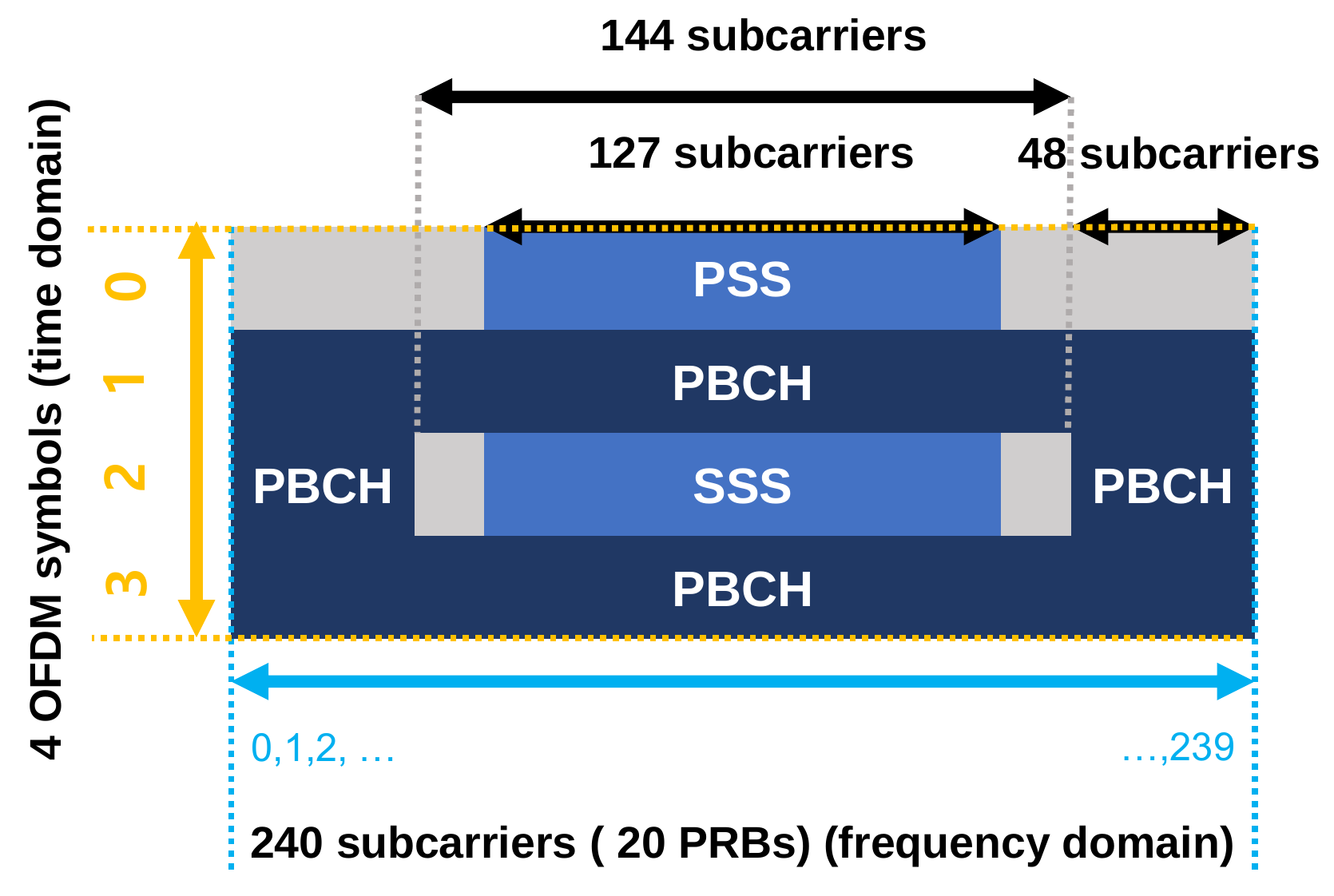}
\caption{SSB Configuration}%
\label{SSB}
\end{figure}

The structure of the SSB is illustrated in Fig.\ref{SSB}.
In a time domain perspective, the SSB occupies 4 Orthogonal Frequency Division Multiplexing (OFDM) symbols. The first and third OFDM symbols include the PSS and the SSS, respectively, and the last 3 OFDM symbols contain the PBCH. 
In a frequency domain perspective, the SSB occupies 240 subcarriers, which correspond to 20 Resource Blocks (RBs). 
The gNB modulates downlink OFDM symbols with one of the candidate SubCarrier Spacing (SCSs) parameters referred in \cite{TS38211}.
The SSB also has its own index and a set of SSBs having distinct indices are successively broadcasted in a period.
Each SSB is sent over a distinct beam and at a distinct moment, so that the UE can identify the beam by the SSB index.
Fig.\ref{fig_1} shows an example of the SSB burst transmission, where $L_{max}$ SSBs are transmitted in a period.

\begin{table}[htbp]
  \centering
  \caption{SSB Configuration}
    \begin{tabular}{|c|c|c|}
    \toprule
    \multicolumn{1}{|l|}{channel or signal} & \multicolumn{1}{l|}{OFDM symbol Indices} & \multicolumn{1}{l|}{Subcarrier Indices} \\
    \midrule
    PSS   & 0     & 56,57,...,182 \\
    \midrule
    SSS   & 2     & 56,57,...,182 \\
    \midrule
    \multirow{2}[4]{*}{Set to 0} & 0     & 0,1,2,...55,183,184,...,239 \\
\cmidrule{2-3}          & 2     & 48,49,...,55,183,184,...,191 \\
    \midrule
    \multirow{2}[4]{*}{PBCH} & 1,3   & 0,1,...,239 \\
\cmidrule{2-3}          & 2     & 0,1,...,47, 192,193,...,239 \\
    \midrule
    \multirow{3}[4]{*}{DMRS} & 1,3   & 0+v, 4+v, 8+v, ..., 236+v \\
\cmidrule{2-3}          & \multirow{2}[2]{*}{2} & 0+v, 4+v, 8+v, ..., 44+v \\
          &       & 192+v, 196+v, ...,236+v \\
    \bottomrule
    \end{tabular}%
  \label{SSBConfig}%
\end{table}%

Table.\ref{SSBConfig} describes how the PSS, SSS and PBCH are multiplexed in an SSB in terms of the OFDM symbol and subcarriers indices \cite{TS38211}.
It shows which symbols in an SSB correspond to the PSS, SSS or PBCH.
For instance, PSS occupies the center subcarriers of the first OFDM symbol, so the OFDM symbol index is 0 and the subcarrier indices are from 56 to 182.
Also, notable thing is that the PBCH consists of the symbols for a Broadcast CHannel (BCH) transport block and DeModulated Reference Signal (DMRS).
Here, the subcarrier indices for the DMRS depend on $v$, which is derived as modulo 4 of the $N_{ID}^{cell}$.
The DMRS symbols can be used as received pilot signal for channel estimation and consequent channel compensation for the BCH transport block.
Using the zero-forcing scheme, the channel $H$ can be derived as following,
\begin{equation}
H=X_p^* Y_p,
\label{chComp1}
\end{equation}
where $(\cdot)^*$ denotes conjugation, and $X_p$ and $Y_p$ denote the original DMRS symbol sent by the gNB and the received DMRS symbol, respectively. A received symbol of the BCH transport block $Y_d$ can be compensated to $\tilde{X_d}$ as following,
\begin{equation}
\tilde{X_d} = \frac{Y_d}{H} = \frac{Y_d H^{*}}{||H||^{2}}.
\label{chComp2}
\end{equation}
This one-tap equalization sufficiently compensates the phase distortion by the channel and is essential for the UE to achieve the performance of the further channel decoding process for the PBCH.

The original DMRS sequence $rs$ in the PBCH depends on the SSB index, denoted by $i_{SSB}$, as well as $N_{ID}^{cell}$. Each symbol in a DMRS sequence is defined in \cite{TS38211} as following,
\begin{equation}
rs[m]={1\over{\sqrt2}}(1-2c[2m])+j{1\over{\sqrt2}}(1-2c[2m+1]).
\end{equation}
c[n] is a generic pseudo-random sequences defined by a length-31 gold sequence as following, 
\begin{equation}
c[n]=(x_{1}(n+N_{c})+x_{2}(n+N_{c}))\!\!\!\!\!\!\mod\!\!2, 
\end{equation}
\begin{equation}
x_{1}(n+31)=(x_{1}(n+3)+x_{1}(n))\!\!\!\!\!\!\mod\!\!2,
\end{equation}
\begin{equation}
x_{2}(n+31)=(x_{2}(n+3)+x_{2}(n+2)+x_{2}(n+1)+x_{2}(n))\!\!\!\!\!\!\mod\!\!2,
\end{equation}
where $N_{c}$=1600. 
$x_{1}$(n) is initialized with $x_{1}$(0)=1 and $x_{1}$(n)=0 for n=1,2,...,30,
and $x_{2}$ is initialized by
\begin{equation}
c_{init}\!\!=\!\!2^{11}(i_{SSB}\!+\!1)(\left[\frac{N_{ID}^{cell}}{4}\right]\!+\!1)+2^6(i_{SSB}\!+\!1)+N_{ID}^{cell}\!\!\!\!\!\!\mod\!4. 
\end{equation}

Consequently, the SSBs sent through distinct beams contain different DMRS sequences, and the UE needs to choose a proper DMRS sequence which corresponds to the received beam before the channel compensation procedure.

Here, we will focus on the main problem to find out the SSB index from the received signal.
At the moment of the PBCH decoding, the UE can derive $N_{ID}^{cell}$ from the PSS and SSS detections, but does not have any prior information about the SSB index. 
Thus, the UE needs to blindly figure out the SSB index and the original DMRS sequence for the channel estimation.
We will further assume that the UE has successfully detected PSS and SSS for a certain SSB and recognizes $N_{ID}^{cell}$. 
From received signal, the UE needs to find the best $i_{SSB}$ whose DMRS sequence $rs$ is the most suitable among the candidate DMRS sequences.

\subsection{Correlation-Based Blind Detection}
One straight-forward way to solve the problem is to use correlation formula, which can let us know how much a target sequence is similar with an original sequence that we look for. 
As the UE has found out $N_{ID}^{cell}$ from PSS and SSS detections, it can generate $L_{max}$ candidate DMRS sequences by (4). 
By comparing the correlations of the received DMRS sequence and the candidate DMRS sequences, the UE can simply select one DMRS sequence which is the most likely to be similar with the received DMRS sequence.

In an Additive White Gaussian Noise (AWGN) channel environment, finding the best suitable one from candidate DMRS sequences is equivalent to the problem of the maximum likelihood decision. 
The maximum likelihood decision finds the DMRS sequence with the shortest Euclidean distance.  
Let $d^{2}(x,y)$ denotes as the Euclidean distance between signal x and y and $s_{i}$ denotes as i'th candidate DMRS sequence.
In a mathematical perspective, the maximum likelihood decision can be expressed as following,  
\begin{equation}
i= \arg\min_i d^{2}(\boldsymbol{rF_{h}}, \boldsymbol{s_{i}})
\end{equation} 
\begin{equation}
= \arg\min_i \sum_{k} \parallel rF_{h}[k]-s_{i}[k]\parallel ^{2}
\end{equation} 
\begin{equation}
\label{MLE}
= \arg\min_i \sum_{k} \parallel \!\!rF_{h}[k]\!\!\parallel ^{2} + \parallel \!\!s_{i}[k]\!\!\parallel^{2} -2 Re\{rF_{h}[k] s_{i}^{*}[k] \},
\end{equation} 
where $rF_{h}$ is a received DMRS sequence.  
A DMRS sequence is composed of QPSK symbols with constant symbol energy, so (11) can be simplified as,
\begin{equation}
i = \arg\max_i Re\{\boldsymbol{rF_{h}} \cdot \boldsymbol{s_{i}}\},
\end{equation} 
where ($\cdot$) denotes inner product. Using $N_{ID}^{cell}$ derived from the PSS and SSS detections, UE can generate $s_{i}$ for each $i$ and derive the inner product result in (12).
Then, the UE can select the best $i$ which maximizes the inner product result and regard it as the SSB index.

This correlation-based blind detection is advantageous in view of simple implementation and reasonable computational complexity, which is the reason why we selected it as a base-line scheme.
However, this scheme may frequently cause false detections when SNR is low and the effect of noise becomes significant. 
This causes to degrade the performance of the PBCH decoding, since it misleads the channel compensation process.

\section{Blind SSB Index Detection Using Channel-Learning Models}
\begin{figure}
\includegraphics[width=\columnwidth]{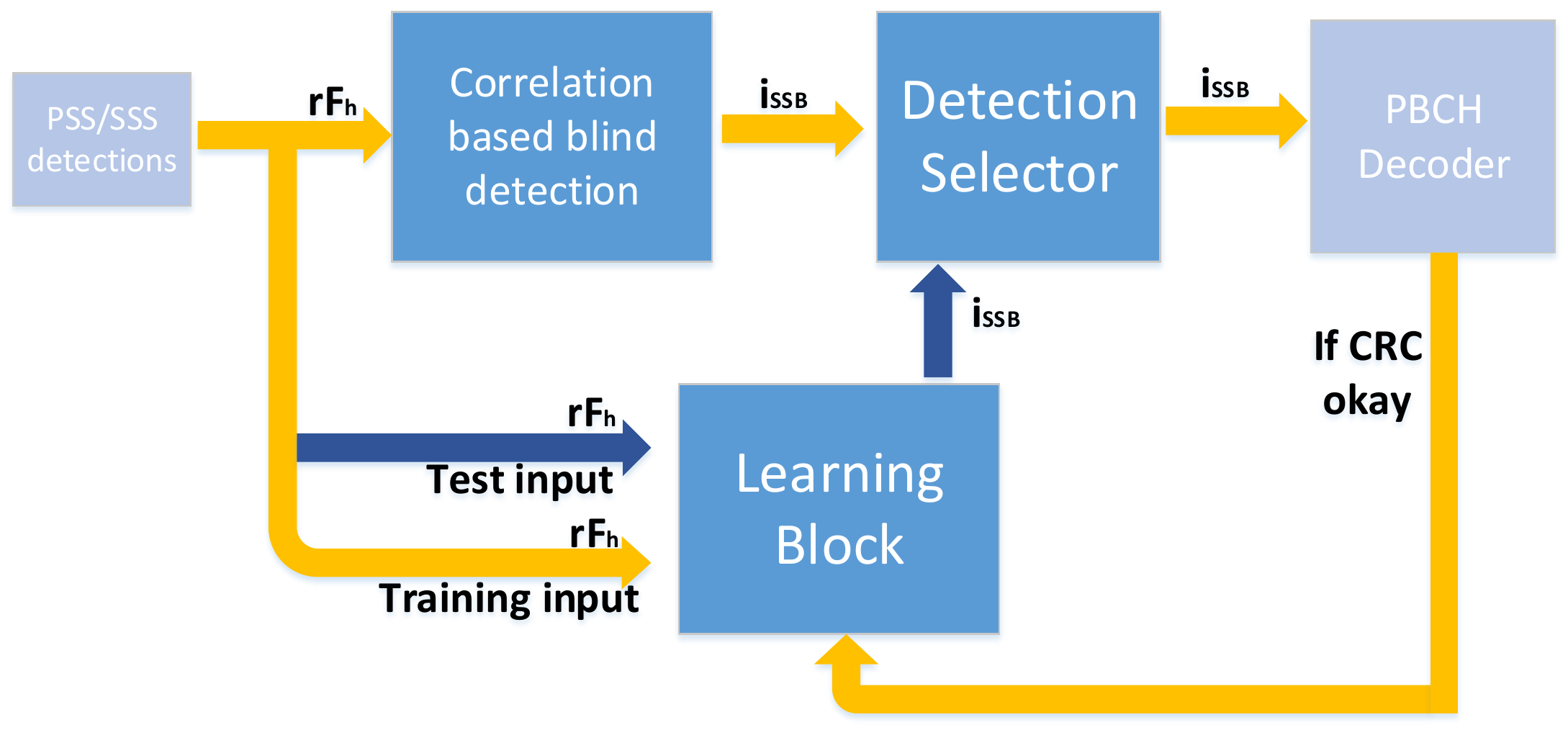}
\caption{The block diagram of the proposed scheme}
\label{fig_3}
\end{figure}
What is noteworthy in our proposed scheme is to statistically predict the current channel through the accumulated data of the past channel. 
The correlation scheme does not consider the statistical characteristics of the channel since it finds the $i_{SSB}$ based on the current received signal only. 
The proposed scheme in contrast utilizes the previous channel information as well as the current received signal to find the $i_{SSB}$.

Fig.\ref{fig_3} shows the block diagram of the blind detection schemes that we suggest. 
In our proposed schemes, the correlation-based blind detection coexists with the ML-based blind detection, which chooses an SSB index based on channel-learning models.
The learning block in the Fig.\ref{fig_3} refers to the ML-based blind detection and operates based on a channel-learning model.
This parallel structure comes from the fact that the channel-learning models require a certain time of training before they properly find the $i_{SSB}$ by themselves. 
When the proposed schemes initially operate or wireless channel has just changed drastically, the channel-learning models need to be trained until the detection probability becomes to be sufficiently high. 
Until the channel-learning models are sufficiently trained, the proposed schemes need to detect the SSB index by the correlation-based blind detection. 


The learning block provides reliable outputs once the learning block is sufficiently trained, but can be frequently erroneous before the sufficient training. 
The detection selector thus estimates how much the learning block is trained and selects one from the outputs of the correlation-based blind detection and the learning block.
It monitors successful detection rate of the learning block by tracking its outputs.
It will choose outputs from the learning block if the successful rate gets larger than a criterion. 
Otherwise, it will regard that the learning block is not sufficiently trained and select outputs of the correlation-based blind detection.

For realizing the learning block, we choose various types of ML models in order to see how ML models tend to work with the beam index detection. 
In general, ML models can be categorized into two models; supervised and unsupervised learning models. 
The supervised learning model has the information of correct answers, meanwhile the unsupervised learning model does not have the correct answers.
In our case, we have the correct answer of the SSB index provided by the PBCH decoder and as the answer is a discrete value, it is a classification problem. 
After correlation-based blind detection and PBCH decoding, we can notice if these processes were successful by Cyclic Redundancy Check (CRC). 
The CRC result tells whether SSB index derived from the correlation-based blind detection was correct or not. 
If the CRC result is positive, the derived SSB index can be used as the correct answer of the training input.

Among the supervised learning models, we select the most widely-used classification models for realizing the learning block; Multi Layer Perceptron (MLP) classifier, Logistic Regression, Support Vector Classifer (SVC), Random Forest classifier and Ensemble Voting classifier. 
These five ML models have different approaches to learn channel and detect beam indices. 
MLP classifier utilizes a neural network model with hidden layers and Logistic Regression utilizes a linear model with linear form of decision boundaries. 
SVC classifies by finding maximum margins between data sets. 
Random Forest and Ensemble Voting classifiers are Ensemble-based methods which train each of the learning models and find the best one. 
Observing how these models detect beam indices provides intuitions of how the other similar models will perform with respect to the beam index detection. 
For instance, if the MLP classifier performs the best, it gives an insight that a deep learning model based on a neural network will also perform well. 
If the Logistic Regression performs the worst, then we can assume that other linear models will not also fit well with the beam index detection.

\subsection{Pre Processing for Channel Learning}

Before detecting the SSB index blindly, we need to pre-process the received signal for learning the channel.
The received signal is sampled with a rate of $f_s$ and proceeds cell search for the samples buffered for $T_{bu}$ seconds.
One cell search procedure is consequently done per a sampling data set of $N_s = f_s T_{bu}$ samples.
We denote $r[n]$ $(n = 0, 1, ..., N_s-1)$ as the received signal after the sampling process.
If the gNB is configured to use the SCS $f_{SCS}$, the SSB can only be detected properly with configuring the Fast Fourier Transform (FFT) size $N_{FFT}$ as $f_s/f_{SCS}$.

As we succeed in detecting the PSS and SSS, we can figure out the boundary timing of the SSB, denoted by $n_{ssb}$, and the $N_{cell}^{ID}$ of the gNB. 
Then the sequences of the PBCH symbols are obtained for the further PBCH decoding process.
The three frequency-domain symbol sequences, denoted by $rF_1$, $rF_2$ and $rF_3$, are derived from the following FFT processes,
\begin{equation}
rF_i = FFT(n_{ssb}+n_{cp}+(n_{cp}+N_{FFT})*(i-1)),
\label{rF}
\end{equation}
where $i \in \{1, 2, 3\}$, $FFT(n)$ denotes the FFT result of $N_{FFT}$ consecutive samples starting from $r[n]$, and $n_{cp}$ denotes the number of samples for cyclic prefix. 

To obtain the input features of the learning block, the symbols for the DMRS are extracted from $rF_i$ sequences according to the indices in Table.\ref{SSBConfig}.
The symbol sequence of the DMRS, denoted by $rF_h[f], f=0,1,...,143$, can be extracted as
\begin{equation}
rF_h[m+M_{offset}[i_s] ] = rF_{i} [DMRS_{i}[m] ],
\label{rFh}
\end{equation}
where $i_s \in \{1, 2, 3\}$, $M_{offset} \in \{0, 60, 84 \}$ and $DMRS_i$ is the sequence of the subcarrier indices for the DMRS symbols $rF_{i}$.
This $rF_h$ can be regarded as the input of the learning block and the index of the original DMRS sequence is the target that the learning block needs to derive.

The extracted DMRS sequence $rF_h$ is basically used for the correlation-based blind detection, and is also utilized for training the channel-learning models. 
The operand part of the correlation-based blind detection is composed of 288 real values, which are denoted by $x_{i}, i=1,2,...,288$, and are the real and imaginary parts of $rF_h$. 
This $x_{i}$ is regarded as the input features of the channel-learning models and collected to a storage for the further training process.

To enhance the performance of channel-learning models, normalization is proceeded for input data before training. 
The normalization aims to changes the scale of input data without distorting the differences of the data values.
Input data is normalized with normalization factor $N_{p}$, which can be derived as following, 
\begin{equation}
N_{p} = E[P] = E[ \sum_{i=1}^{288}\parallel \!\!x_{i} \!\!\parallel ^{2}\!\!/288],
\end{equation}
where $E[\cdot]$ denotes the Ensemble expectation for whole data sets.
This normalization process can prevent channel-learning models from adapting to different power levels of input data sets.

\subsection{Channel Learning Model I : Multi Layer Perceptron classifier}

MLP classifier is one of the neural network models which is comprised of input, output and hidden layers. 
The neural network model learns based on the following function, 
\begin{equation}
f(x): R^{n} \to R^{o},
\end{equation}
where n is the number of input features and o is the number of candidate outputs. 
In our case, n is 288 and o is $L_{max}$. 
Between the input and output layers, there are hidden layers and each hidden layer is composed of $a$ neurons. 
Let $a_{k}^{l}$ denote the value of the k'th neuron in l'th hidden layer. 
Each layer is connected to the next layer via weight values.
Let $\textbf{W}^{(l)}$ denotes the weight values mapping from layer (l) to layer (l+1).  
$\textbf{W}^{(l)}=[\textbf{W}_{1}^{(l)}, ..., \textbf{W}_{a}^{(l)}]$, where $\textbf{W}_{k}^{(l)}$ is a vector and constructs $a_{k}^{l+1}$. 
During training, each weight value is updated until the whole weight values are optimized. 

When input data come in, the values of neurons are calculated from the first layer to the output layer, which is called forward propagation.
For the input vector \textbf{X}=$[x_{0}, x_{1}, x_{2}, x_{3}, ..., x_{n}]$ where $x_{0}$ is a bias unit, it is multiplied by $\textbf{W}^{(1)}$ and passed to an activation function $h(x)$. 
This induces the next neuron values. 
For example, the first neuron in layer 2 is calculated by the following equation, 
\begin{equation}
a^{(2)}_{1}=h(\textbf{W_{1}}^{(1)}\textbf{X}^{T}).  
\end{equation} 
When the neurons in the output layer are induced, the SSB index can be selected as the index of the largest neuron in the output layer as follows, 
\begin{equation}
i_{SSB} = \arg\max_ih( \textbf{W}_{i}^{(l)}\textbf{a}^{(l)}), i=0,1,...,L_{max}\!\!-\!1.
\end{equation}
 
\begin{figure}
\includegraphics[width=\columnwidth]{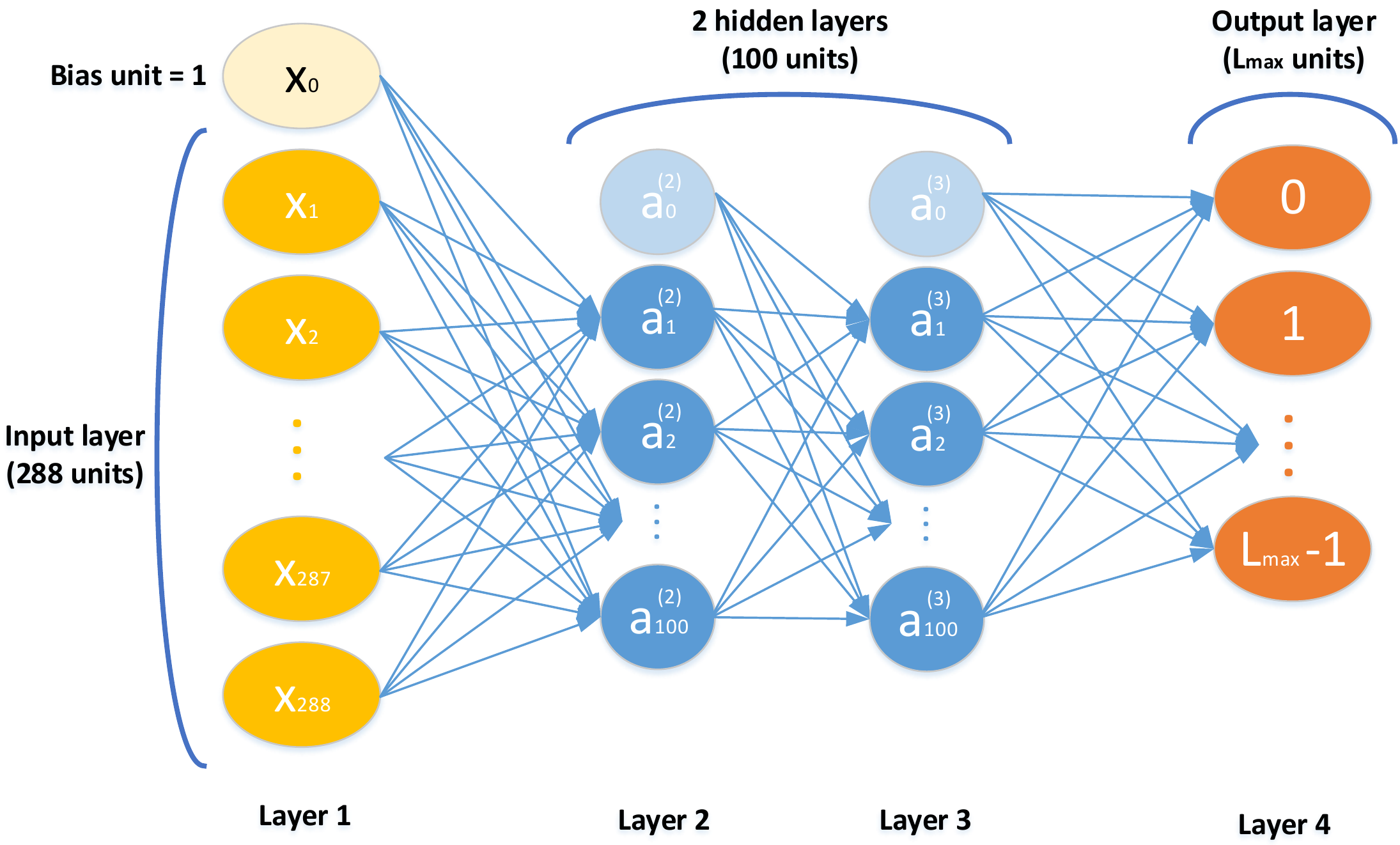}
\caption{The neural network model for the MLP classifier}
\label{fig_4}
\end{figure}

As shown in the Fig.\ref{fig_4}, we design the neural network model for the blind beam index detection with two hidden layers and each layer is composed of 100 neurons. 
To find the optimized weight values, the stochastic gradient-based optimizer $adam$ \cite{adam} is used.
In addition, we use Rectified Linear Unit (ReLU) function $h(x) = max(0, x)$ for the activation function, which returns the input value directly if it is positive and returns zero otherwise. 
We select this activation function because it reduce a gradient vanishing problem \cite{vanishing} which can often happen while optimizing the weight value. The ReLU is also good in a complexity aspect. 
Unlike the hyperbolic tan function ${e^{2ax}-1}\over{e^{2ax}+1}$ and sigmoid function $1\over{1+e^{-x}}$, the ReLU is much simpler and easy to be implemented.

\subsection{Channel Learning Model II : Logistic Regression} 
Logistic Regression is frequently used for classification problems. 
After the input goes to the hypothesis, it is classified as 1 if the output value is greater than 0.5, or 0 otherwise.
Mathematically, the output value $O_{LR}$ can be expressed as,  
\begin{equation}
O_{LR} =
\begin{cases}
0 & \text{if  $h_{\theta}(\boldsymbol{\theta}\boldsymbol{X}^{T})$ $\le 0.5$} \\
1 & \text{if  $h_{\theta}(\boldsymbol{\theta}\boldsymbol{X}^{T})$ $\ge 0.5$},
\end{cases}
\end{equation}
where $\boldsymbol\theta$=$[\theta_{0}, \theta_{1}, \theta_{2}, \theta_{3}, ..., \theta_{n}]$ and are the parameters of the Logistic Regression.
While training the model, the parameter vector $\boldsymbol{\theta}$ are continuously updated. 
This $\boldsymbol\theta$ is optimized by Limited memory-Broyden Fletcher Goldfarb Shanno (L-BFGS) \cite{LBFGS} optimizer.
For multi-class classification where there are o output candidates, the Logistic Regression model needs o hypotheses 
to classify the output as follows, 
\begin{equation}
h_{\theta}^{(i)}(\boldsymbol{\theta}\boldsymbol{X}^{T})=P(y=i|\theta), i \in \{0,1,2,...,o-1\}.
\end{equation}

For simple calculation, we use the sigmoid function $h_{\theta}(x)={1\over{1+e^{- x}}}$ as the hypothesis function. 
As there are $L_{max}$ output candidates in our case, we can regard this blind detection problem as a multi-class classification, and the output is induced as follows, 
\begin{equation}
i_{SSB}=\arg\max_i{h_{\theta}^{(i)}(\boldsymbol{\theta}\boldsymbol{X}^{T})}, i \in \{0,1,2,...,L_{max}\!-1\}.
\end{equation}

\begin{figure}
\includegraphics[width=\columnwidth]{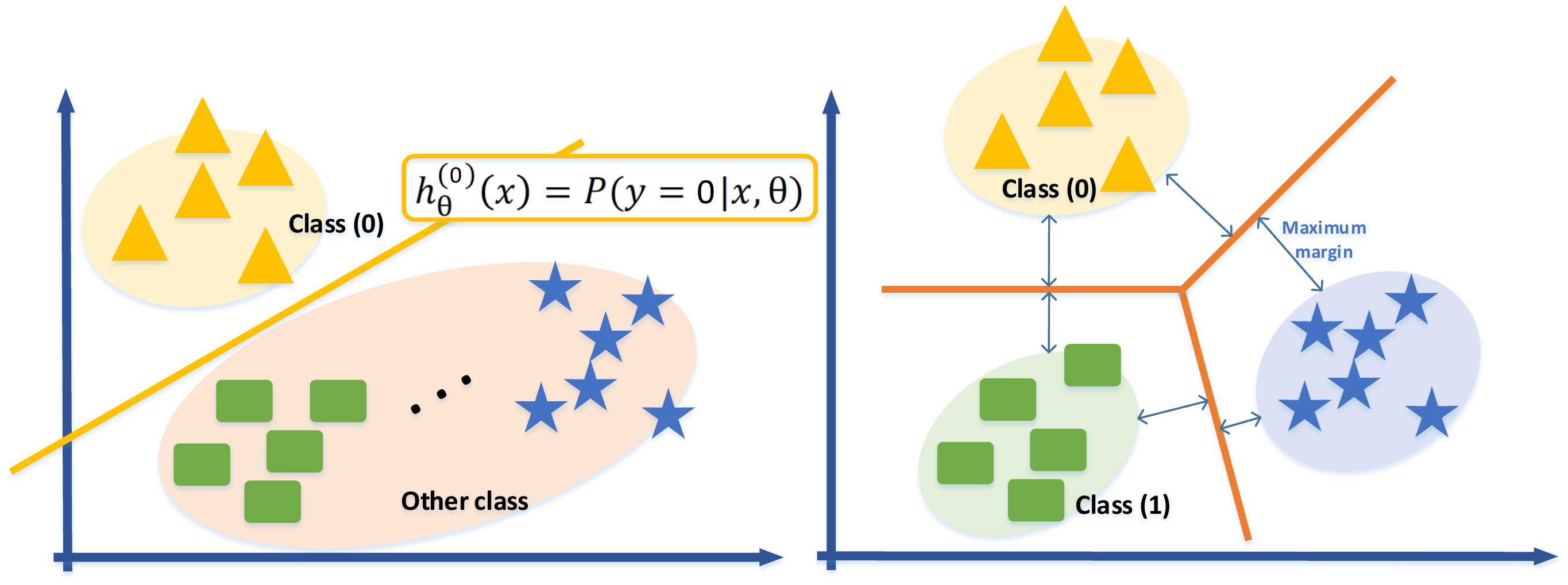}
\caption{An example of the Multi-class classification and SVC}
\label{fig_5}
\end{figure}

\subsection{Channel Learning Model III : Support Vector Machine for Classification}
The SVC refers to a specific version of the Support Vector Machine (SVM) suitable for classification. 
Advanced from the Logistic Regression model, it results in making a decision boundary to have a bigger margin between classes. 
It predicts output value as 1 if $\boldsymbol\theta \textbf{X}^{T} \!>\! 0$ and as 0 if $\boldsymbol\theta \textbf{X}^{T} \!<\! 0$. 

To make a non-linear decision boundary, we can use a concept of kernel in SVC. 
Given m training examples ${(\textbf{X}^{(1)}, Y^{(1)}), (\textbf{X}^{(2)}, Y^{(2)}), ..., (\textbf{X}^{(m)}, Y^{(m)})}$, where $\textbf{X}^{(i)}$ is i'th input training set and $Y^{(i)}$ is its correct answer,
landmark $\textbf{l}^{(i)}$ is determined by
\begin{equation}
\textbf{l}^{(1)}=\textbf{X}^{(1)}, \textbf{l}^{(2)}=\textbf{X}^{(2)}, ..., \textbf{l}^{(m)}=\textbf{X}^{(m)}. 
\end{equation}
Consequently, for training example $(\textbf{X}^{(i)}, Y^{(i)})$, a set of functions are defined as following,
\begin{equation}
\textbf{f}^{(i)}=[f^{(i)}_{0}, f^{(i)}_{1}, f^{(i)}_{2},..., f^{(i)}_{m}].
\end{equation}
\begin{equation}
f^{(i)}_{k}=similarity(\textbf{X}^{(i)}, \textbf{l}^{(k)}),
\end{equation}
where $k=1, 2, ..., m$ and $f^{(i)}_{0}$=1. 
Here, $\textbf{f}^{(i)}$ is decided by the similarity function with an input training sample and landmark.
The SVC with kernel predicts output value by using $\boldsymbol\theta\textbf{f}^{(i)T}$ instead of $\boldsymbol\theta \textbf{X}^{(i)T}$.

In order to make a good performance, we make a non-linear boundary by Radial Basis Function (RBF) \cite{RBF} whose similarity function can be defined as following, 
\begin{equation}
similarity(\textbf{X}^{(i)}, \textbf{l}^{(k)})=exp(-{{|\textbf{X}^{(i)}-\textbf{l}^{(k)}|^{2}}\over{2\sigma^{2}}}). 
\end{equation}
\subsection{Channel Learning Model IV and V : Ensemble-based methods for Classification}
\begin{figure}
\includegraphics[width=\columnwidth]{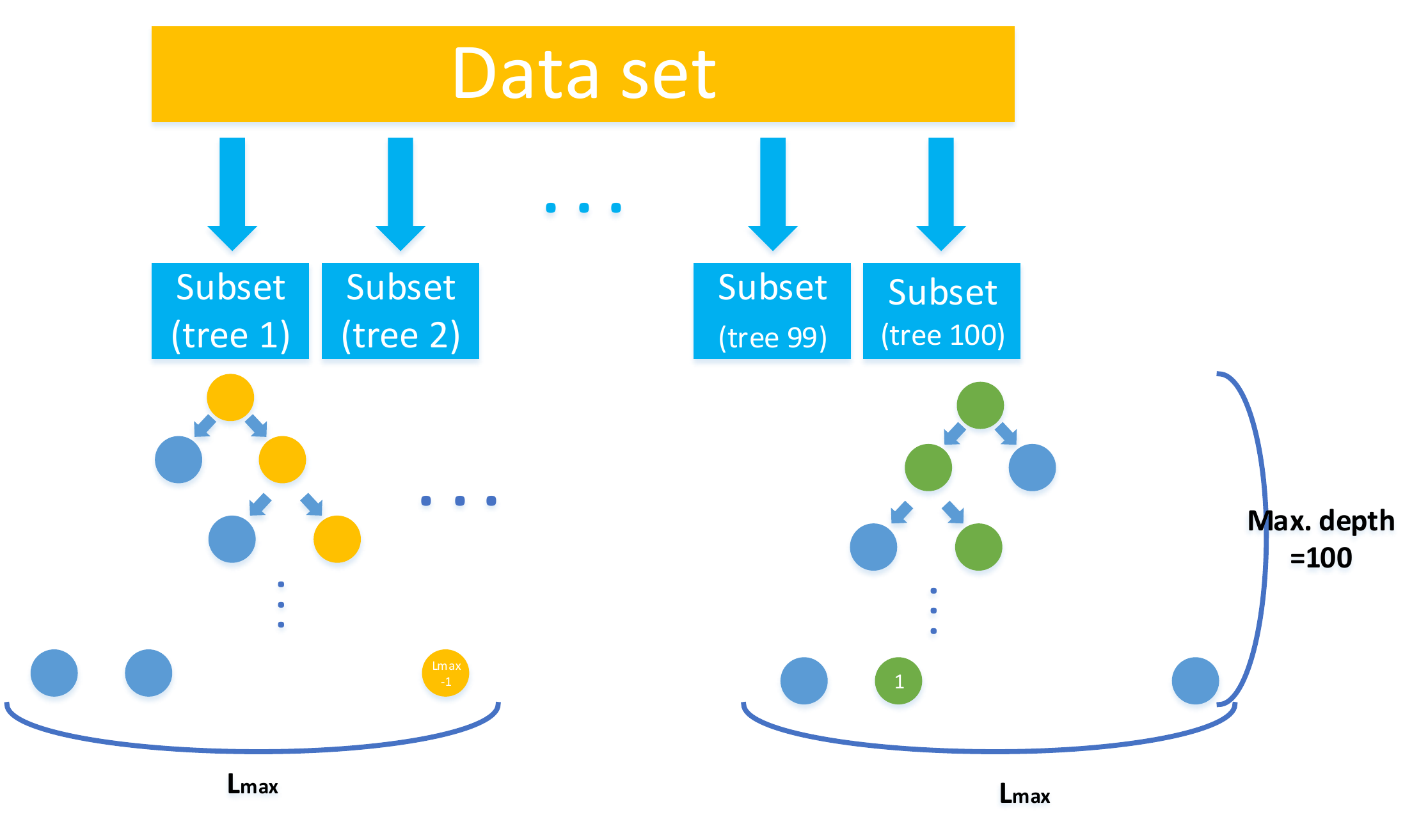}
\caption{The decision tree model of the Random Forest classifier}
\label{fig_7}
\end{figure}

The Random Forest classifier is based on decision trees (subsets). 
The model chooses the output based on the trees most frequently selected. 
When the data comes in, the data is passed to either left or right child node. 
The classification is designed to use the function $gini$ \cite{gini}, which measures impurities at the corresponding node of the tree. 
This Random Forest model classifies the data so that the child nodes have lower impurity than the parent node, and stops when impurity goes down to 0. As in Fig.\ref{fig_7}, we design the number of subsets as 100 and the depth of the tree as 100.  

The Ensemble Voting classifier collects results from each classifier and makes a new output by a majority vote in order to make better classifiers through the model already trained. 


\section{Performance evaluation}
\subsection{Implementation of SDR testbed}
For evaluating the performance of our proposed schemes, we implemented the 5G cell search function on a SDR testbed which operates in real-time.
We used an USRP B210 as an RF signal generator or a receiver and a PC which processes baseband signal.
Once the UE side USRP converts RF signal to baseband signal and passes to the UE PC, the UE PC proceeds the cell search.
The gNB side PC generates the baseband signal of SSBs, and the attached USRP converts it to RF signal and sends to the channel.
We used Open Air Interface open source code \cite{OAI} to implement the gNB on the PC, and our own implementation code of the 5G cell search at the UE side.

\begin{figure}
\includegraphics[width=\columnwidth]{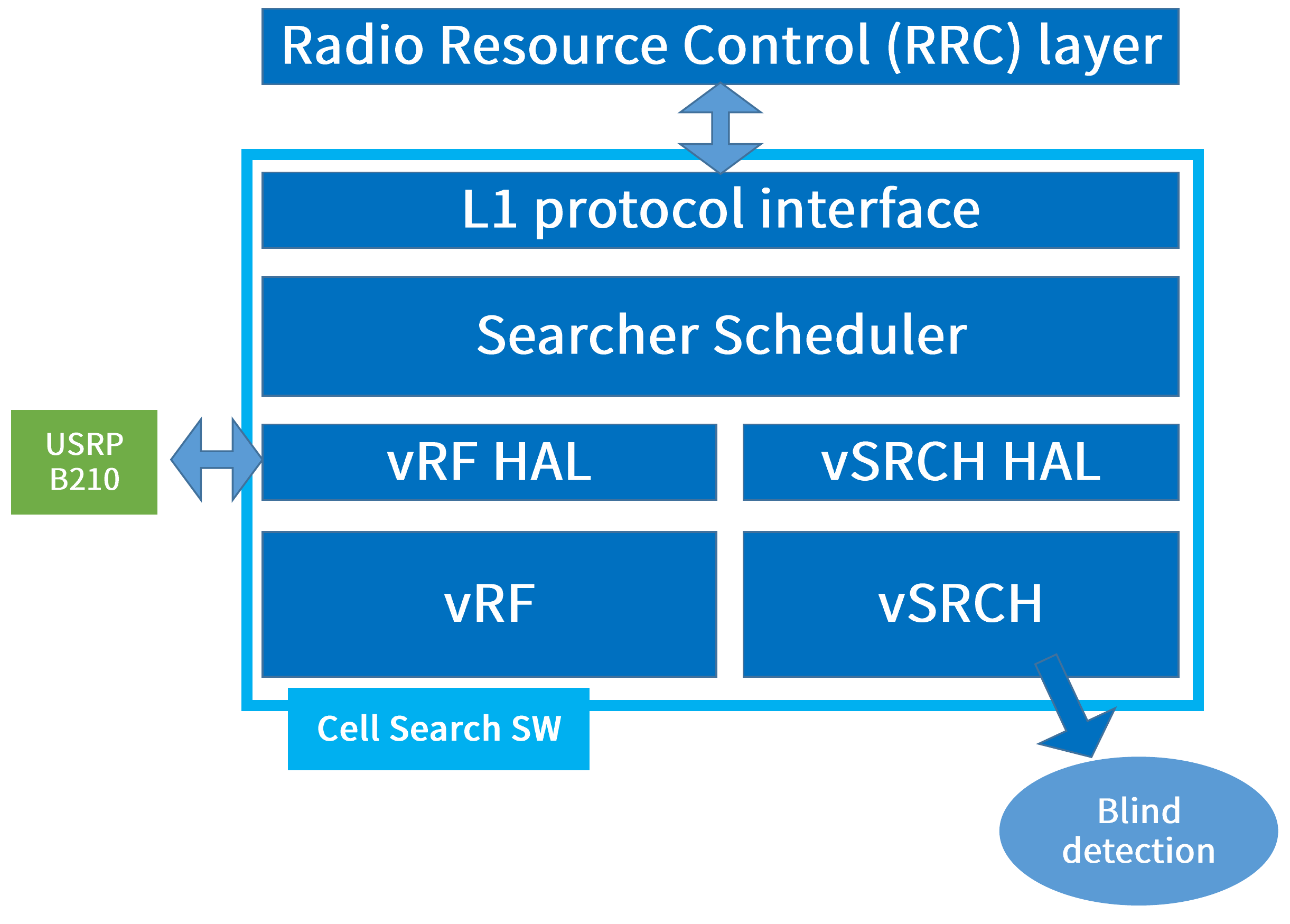}
\caption{The architecture of our cell searcher software}
\label{VHW}
\end{figure}

Fig.\ref{VHW} shows the architecture of our implementation based on an SDR platform. 
Except for the USRP B210 which generates RF signal, the 5G cell searcher is fully softwarized and runs on a general-purpose CPU.
The virtual RF (vRF) and Search (vSRCH) are the software threads that operate as baseband processing hardwares.
The vRF and vSRCH continuously execute RF processing and cell search algorithms, respectively, and deduce search results  cooperatively. 
The functions of the blind detection that we have proposed run in vSRCH. 
The detail specification of the testbed is listed in Table.\ref{hwspec}.
In addition, the parameters of the frame structure, which is shown in Table.\ref{parameter}, are set based on the cell configurations used by SK Telecom 5G gNBs.

\begin{table}[htbp]
  \centering
  \caption{SDR PC specification}
    \begin{tabular}{|cc|}
    \toprule
    \multicolumn{2}{|c|}{CPU: 8th Generation Intel CoreTM i7 Processors} \\
    \midrule
    \multicolumn{1}{|c|}{number of cores} & 6 \\
    \multicolumn{1}{|c|}{number of threads} & 12 \\
    \multicolumn{1}{|c|}{Processor Base Frequency} & 3.70GHz \\
    \midrule
    \multicolumn{2}{|c|}{USRP B210 } \\
    \midrule
    \multicolumn{2}{|c|}{RF coverage from 70MHz-6GHz} \\
    \multicolumn{2}{|c|}{Standard-B USB 3.0 connector} \\
    \multicolumn{2}{|c|}{2 TX \& RX, Half or Full Duplex} \\
    \multicolumn{2}{|c|}{Up to 56 MHz of instantaneous bandwidth in 1x1} \\
    \bottomrule
    \end{tabular}%
  \label{hwspec}%
\end{table}%

\begin{table}[htbp]
  \centering
  \caption{The frame parameters of the test environments}
    \begin{tabular}{|c|c|}
    \toprule
    parameters & value \\
    \midrule
    sampling rate  & 30720000 \\
    Downlink frequency & 3608.79MHz\\
    nr-arfcn & 640586 \\
    Bandwidth & 20MHz\\
    FFT size & 1024 \\
    Subcarrier spacing & 30kHz  \\
    SSB period & 20ms  \\
    $L_{max}$  & 8 \\
    SSB time pattern & case B\\
    \bottomrule
    \end{tabular}%
  \label{parameter}%
\end{table}%

\subsection{Two test scenarios and results}
\begin{figure}
\includegraphics[width=\columnwidth]{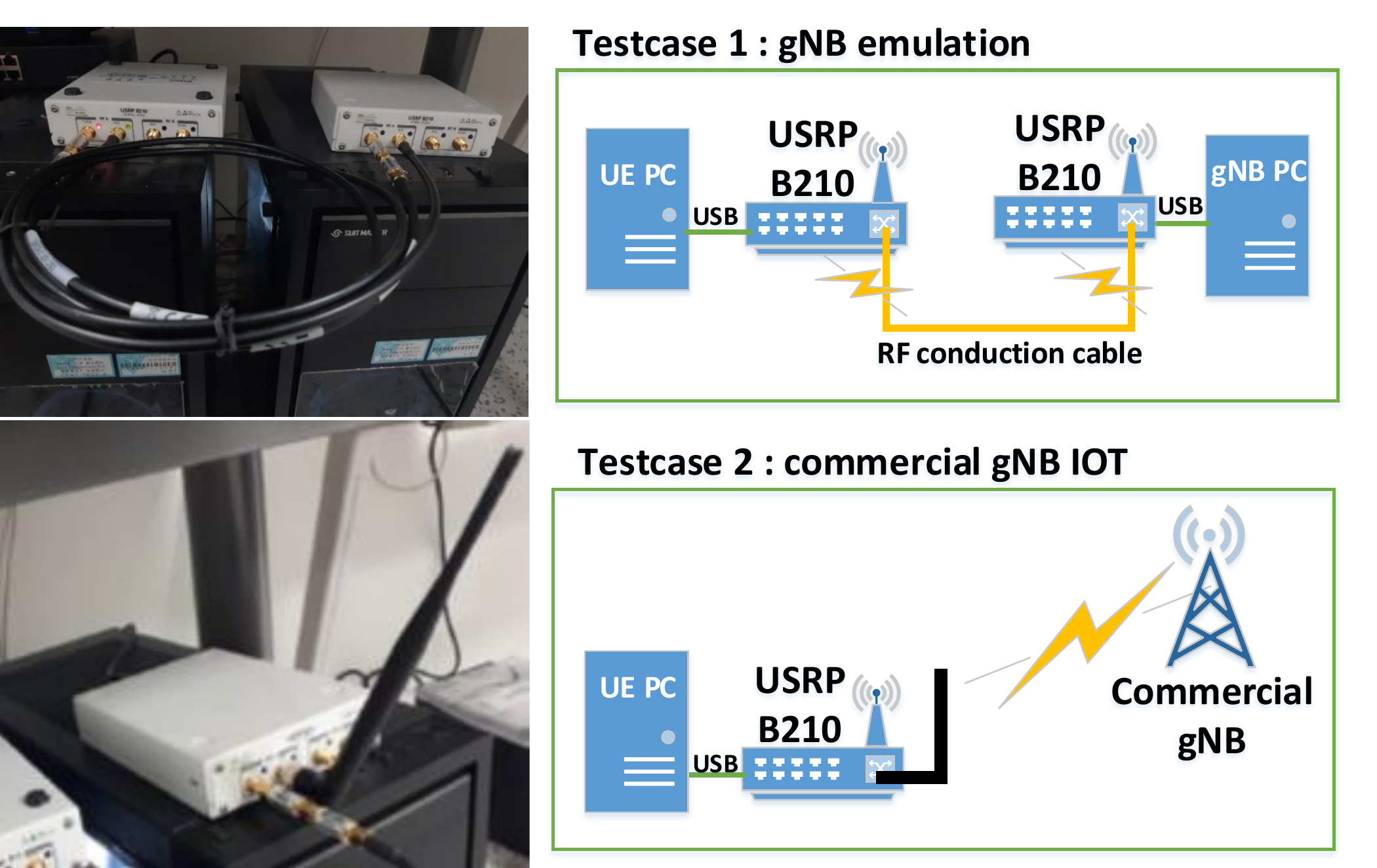}
\caption{The SDR testbed and test scenarios}
\label{tb}
\end{figure}

There are two test scenarios that we have considered; gNB emulation and commercial gNB Inter-Operability Test (IOT). 
In the gNB emulation, the UE side USRP is connected to the gNB side USRP via an RF conduction cable and the UE PC tries to search a cell from the RF signal sent through the cable. 
In this emulation case, UE finds a beam index from a non-directional signal sent through the RF cable.
This test scenario is for observing the performance in a unit-test aspect with varying the received signal environment.
In the commercial gNB IOT, a 3.5GHz antenna is attached to the UE side USRP and the UE PC tries to search a cell from the on-air RF signal of a nearby 5G commercial gNB. 
This experiment was conducted under real, time-varying and NLOS wireless channel, and in a static mobility scenario. 
This test scenario is for verifying how the proposed schemes work in a commercial channel environment. 
Fig.\ref{tb} describes the testbed environment and test scenarios. 

We regard a cell and beam index search trial as a successful event if the cell searcher finds the SSB index correctly and a fail event otherwise.
We compare the proposed schemes with the conventional scheme that detects the SSB index based on correlation.
For evaluating the proposed schemes, we used Scikit Learn python library to realize the channel-learning models. 
For each model, we used parameters as the default values in Scikit Learn hompage except for the ones mentioned in section \uppercase\expandafter{\romannumeral3}. The detailed description for each model is in \cite{SL1} -- \cite{SL5}.
We captured the data samples of the received DMRS sequences $rF_h$s from the testbed and ran the 5 channel-learning models with the captured data for training or detecting $i_{SSB}$. 
The 70\% of the captured data samples, denoted by training sets, are used for training and the rest, denoted by test sets, are used for verifying the success probability. 
For example, out of 10000 DMRS sequences, 7000 sequences are the training sets which are used to train the models, and the rest 3000 sequences are the test sets which are used for the trials. 

\subsubsection{gNB Emulation Results}

In order to figure out the fail/success probability of the blind detection, the test focuses on the procedure after the PSS and SSS detections. 
We therefore measured the fail probability with varying the strength of the DMRS part in the received signal. 
In each test case, cell search trials are repeated with using the same attenuator and keeping the transmit power levels of PSS, SSS and PBCH constantly so that the average SNR of the DMRS is constant.
The fail probability was estimated by averaging the number of failures for the $L_{max}$ cases of using different beam indices. 

\begin{figure}
\includegraphics[width=\columnwidth]{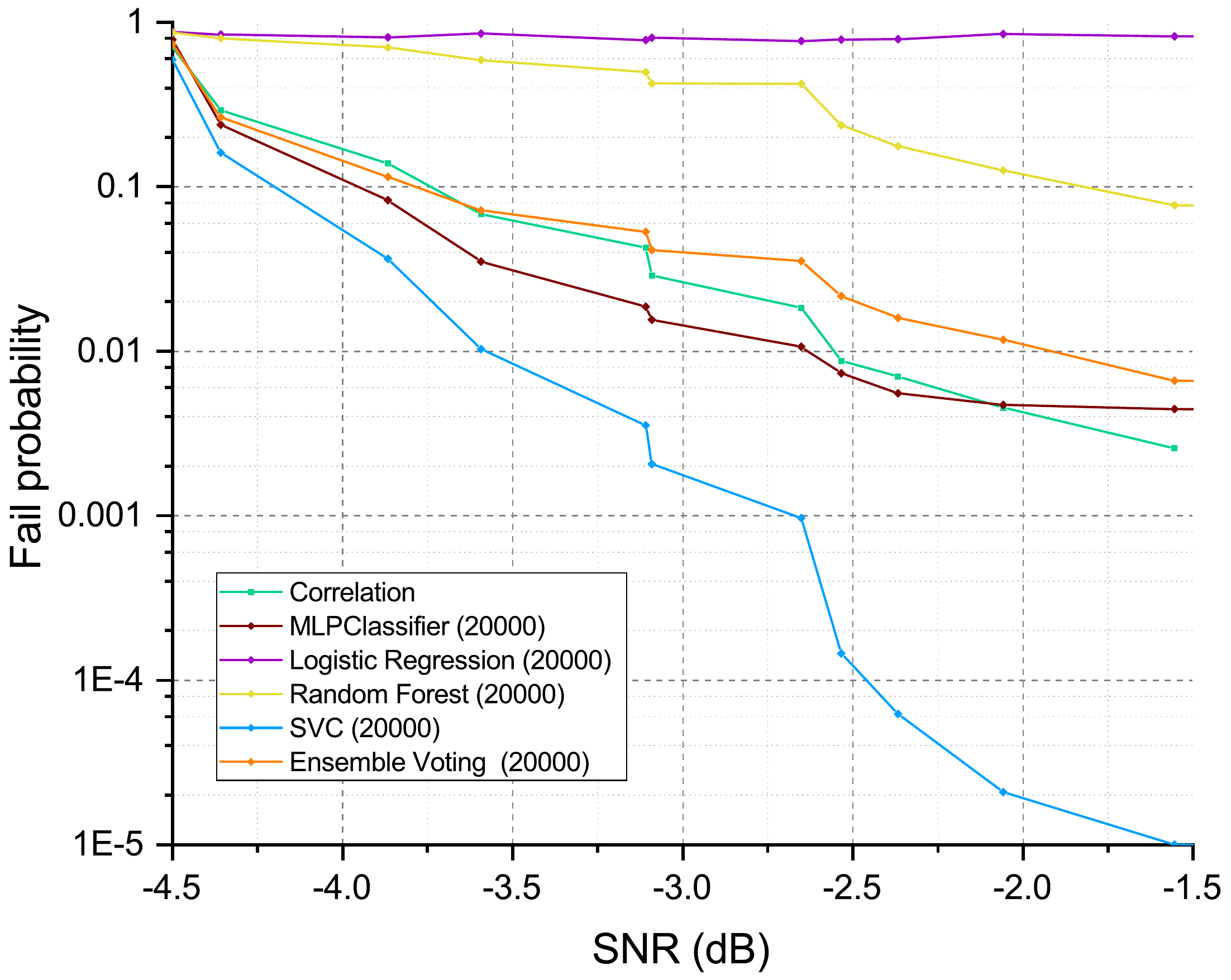}
\caption{The fail probability of the five channel-learning schemes and conventional scheme }
\label{fig_9}
\end{figure}

Fig.\ref{fig_9} shows the results of the fail probability in case of the conventional scheme and the proposed schemes with the five channel-learning models, and trained with 14000 DMRS sequences. 
The results reveal that SVC, MLP classifier and Ensemble Voting classifier have lower fail probability than the conventional scheme in a lower SNR environment. 
The proposed scheme using the Logistic Regression seems to be overfit because the fail probability keeps to be relatively high even when the SNR increases or training is sufficiently done. 
As the Ensemble Voting classifier makes a decision based on the other four channel-learning models, the results show that the fail probability of the Ensemble Voting classifier is between the probabilites of the other channel-learning models. 
In addition, the proposed scheme using the SVC performs with the lowest fail probability among the channel-learning models.
The fail probabilities in case of the Random Forest, MLP and Ensemble Voting classifiers show similar tendencies to the case of the SVC, but are still higher compared with the case of the SVC. 
When the SNR is -3.7dB, the SVC out-performs the MLP classifier, Logistic Regression, Random Forest classifier and Ensemble Voting classifier by 127\%, 2119.6\%, 1820\%, and 214.7\% respectively. 
This guides us that the SVC is the most suitable to the blind detection in the aspect of detection accuracy.

\begin{figure}
\includegraphics[width=\columnwidth]{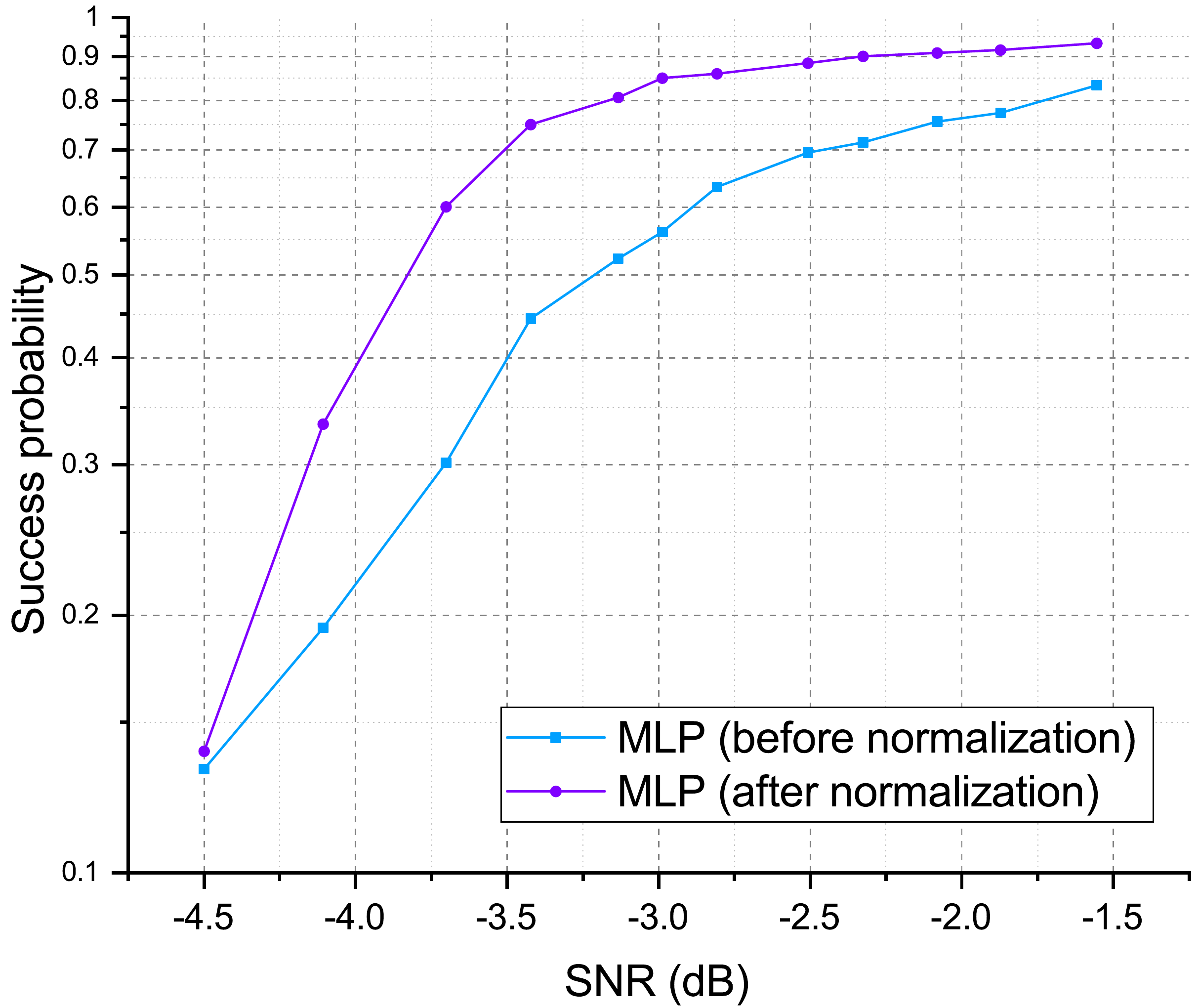}
\caption{The performance of the MLP classifier model trained by 2000 DMRS sequences with and without normaliztion}
\label{MLP_normalize}
\end{figure}

The experimental results also reveal some fundamental facts compatible to the basic ML theory. 
When dealing with multi-dimensions, SVM and neural network tend to perform much better than other channel-learning models \cite{MLbook}.
Here, we have 288 input features so the dimension size of an input is 288. The SVC and MLP classifiers perform better than the other models, which corresponds to the above ML theoretical fact. 
In addition, the neural network is very sensitive to irrelevent features.  
This can be observed by comparing the performance of the MLP classifier with and without normalizaing the input power. 
As shown in Fig.\ref{MLP_normalize}, the normalization enhances the performance by 51.2\% when the SNR is -3dB. 

Compared with the conventional scheme based on correlation, we can see that the proposed schemes can perform better in low SNR cases.
The conventional scheme tends to detect more erroneously when the SNR becomes lower and the fail probability becomes higher than 29\% when the SNR gets lower than -4.1dB.
When the SNR is greater than -2.1dB, the conventional scheme performs with lower fail probability compared with all the channel-learning models except for the SVC.
On the other hand, as the SNR becomes smaller than -2.1dB and -3.5dB, the proposed schemes using the MLP classifier and Ensemble Voting classifier also out-perform the conventional scheme, respectively. 
This reveals that our proposed schemes are effective in a low SNR environment, where various channel-learning models are more preferable to find the original DMRS sequence compared with correlation.

\begin{figure}
\includegraphics[width=\columnwidth]{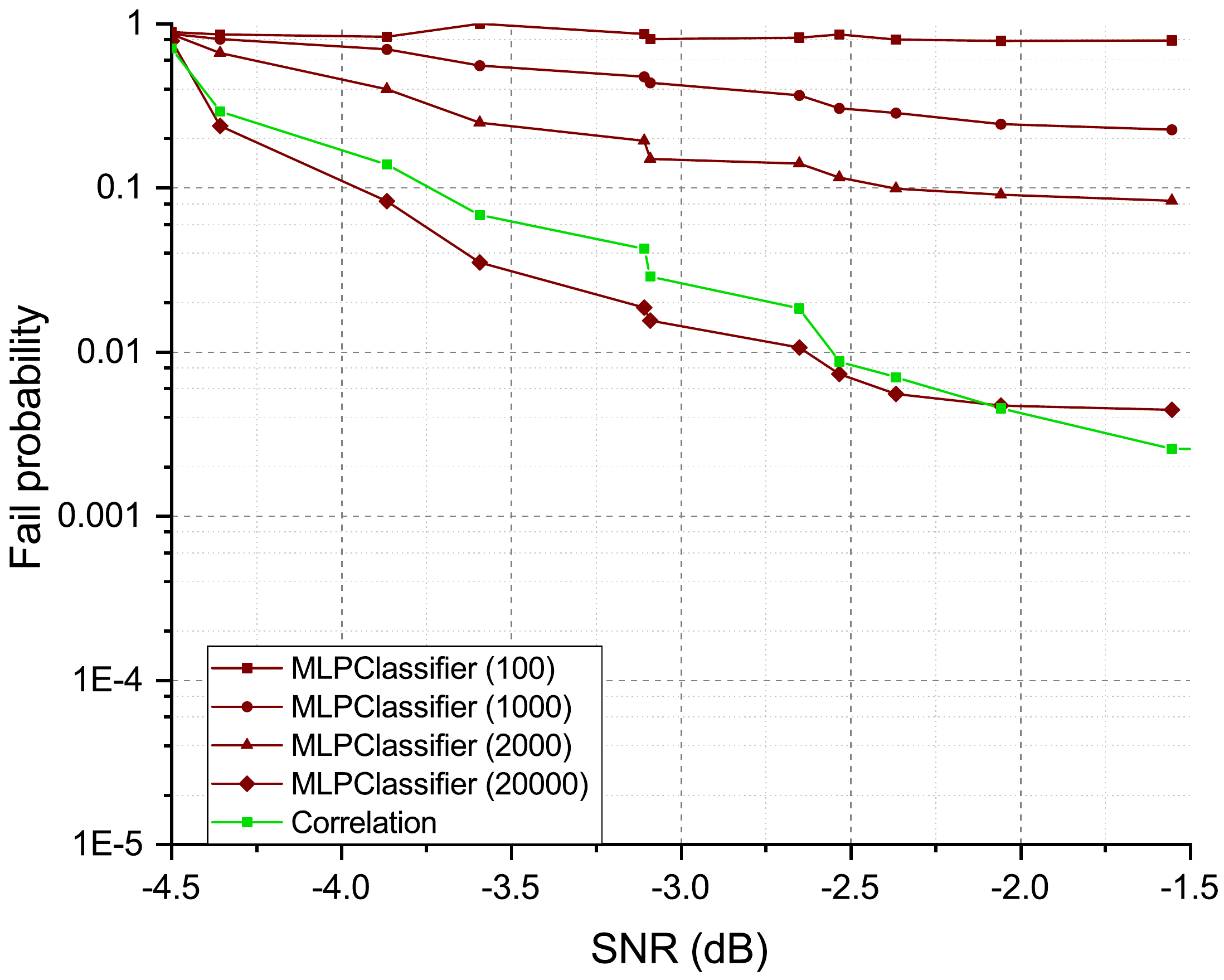}
\caption{The performance of the MLP classifier model}
\label{fig_10}
\end{figure}

\begin{figure}
\includegraphics[width=\columnwidth]{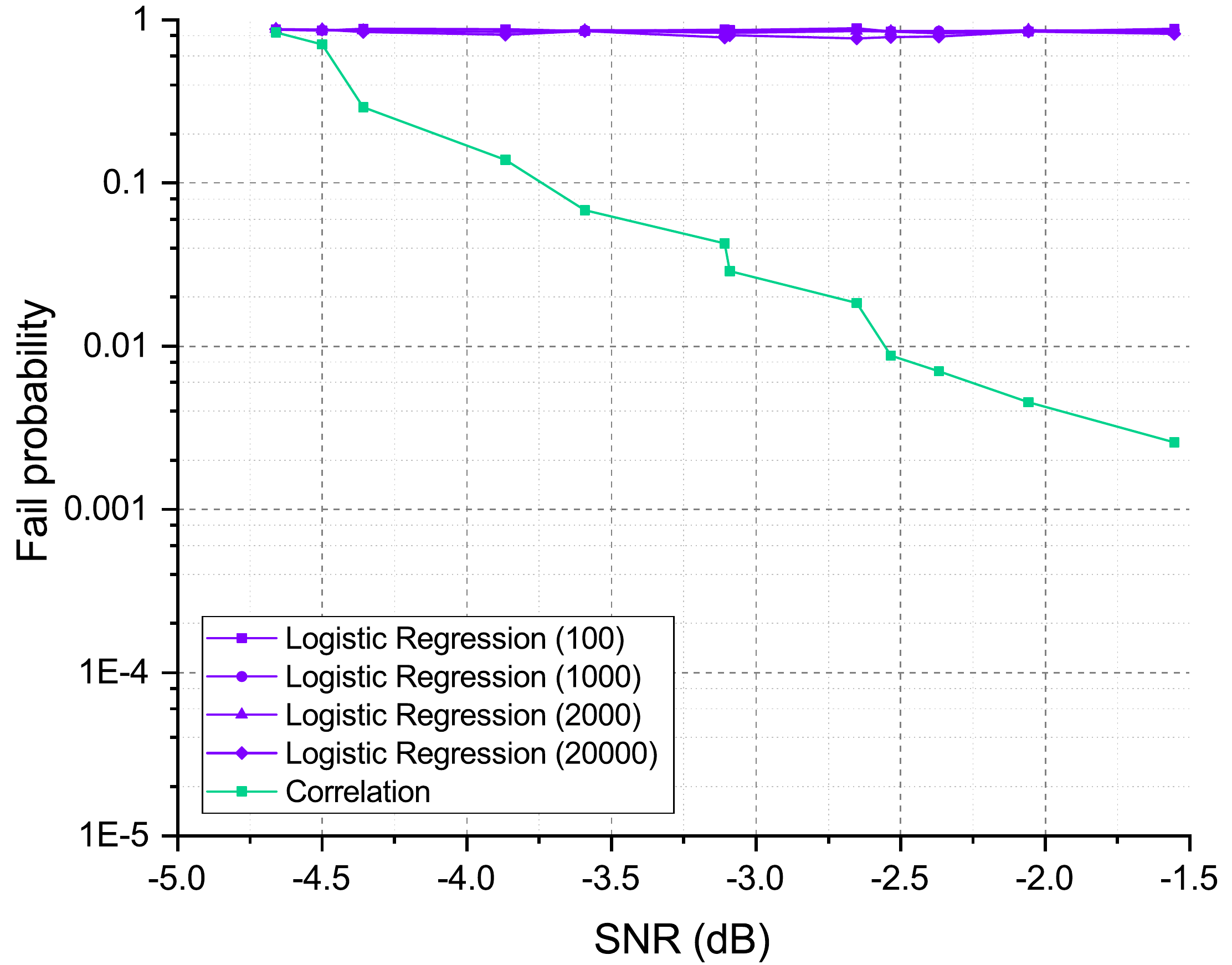}
\caption{The performance of the Logistic Regression model}
\label{fig_11}
\end{figure}

\begin{figure}
\includegraphics[width=\columnwidth]{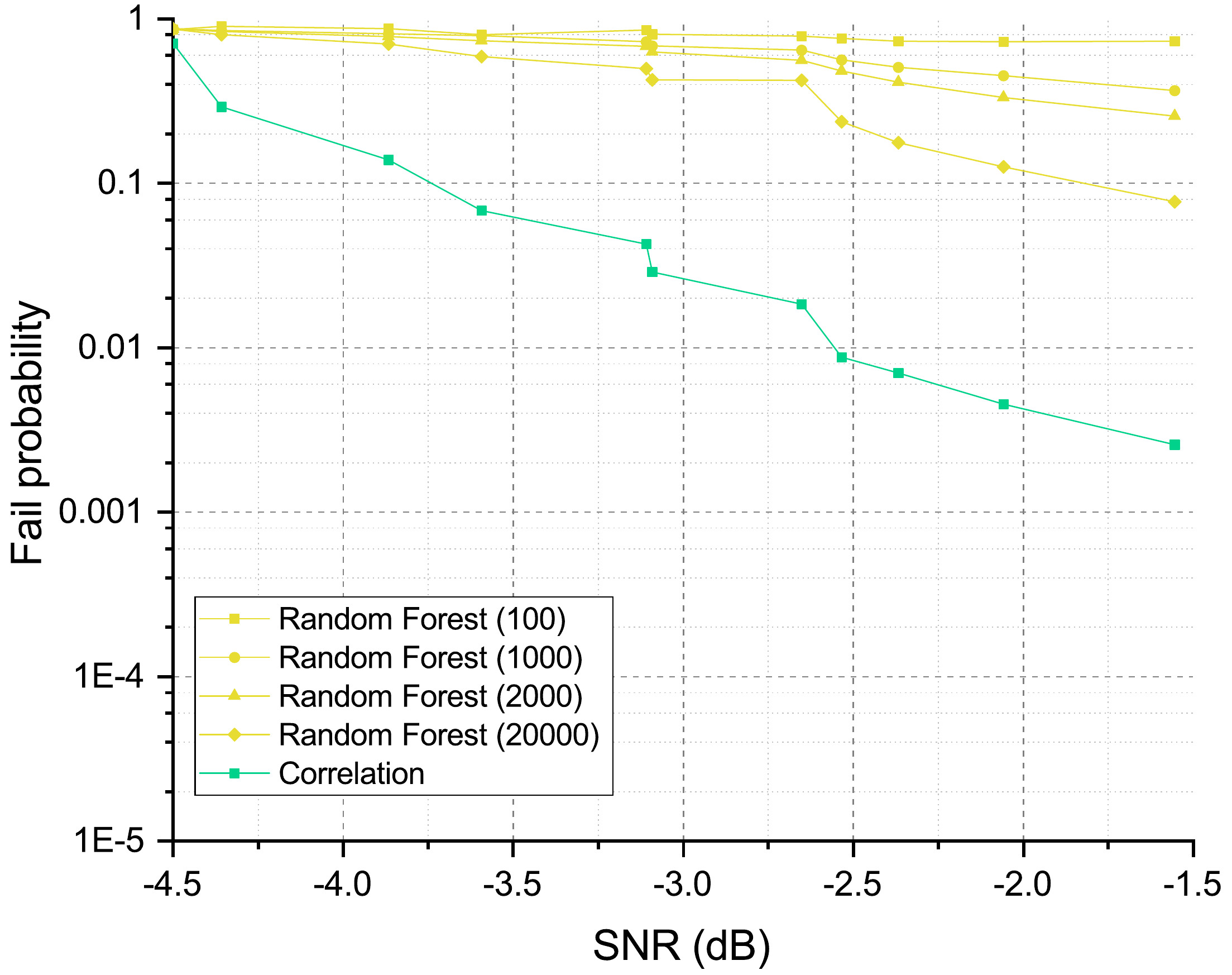}
\caption{The performance of the Random Forest classifier model}
\label{fig_12}
\end{figure}

\begin{figure}
\includegraphics[width=\columnwidth]{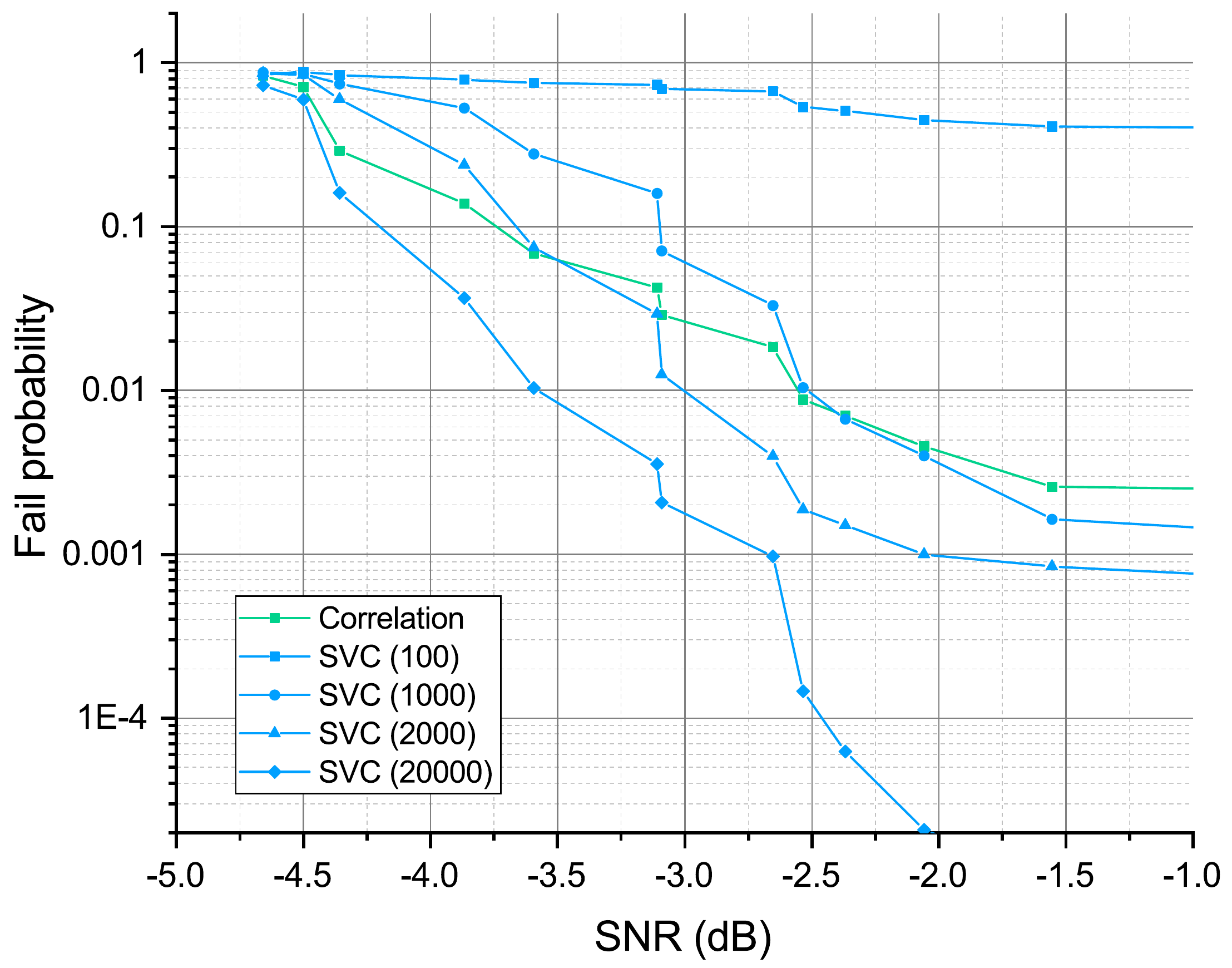}
\caption{The performance of the SVC model}
\label{fig_13}
\end{figure}

\begin{figure}
\includegraphics[width=\columnwidth]{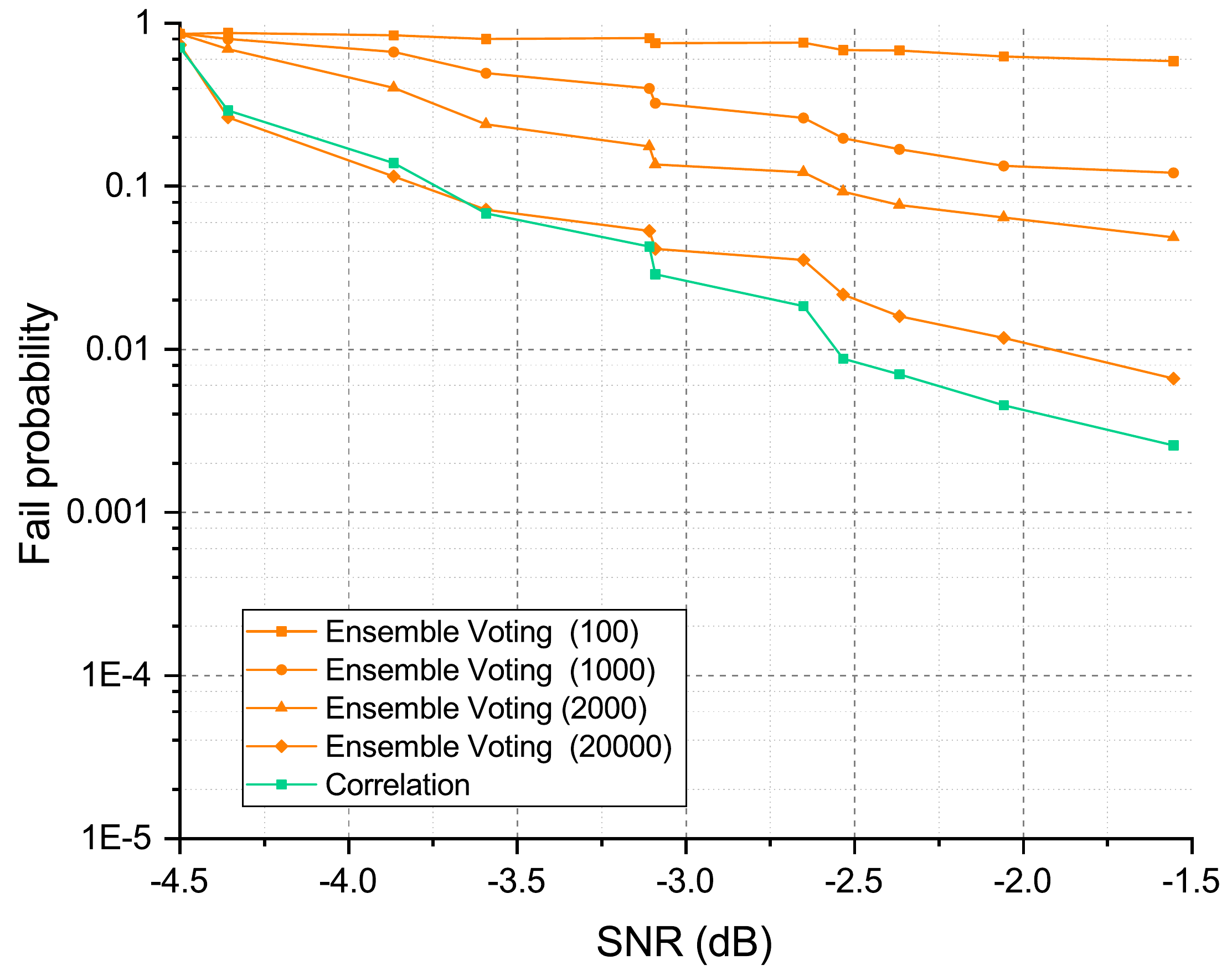}
\caption{The performance of the Ensemble Voting classifier model}
\label{fig_14}
\end{figure}

Fig.\ref{fig_10} -- \ref{fig_14} shows the effect of the number of training by comparing the results of the conventional scheme with each of the five channel-learning models. Each figure includes the fail probability of the conventional and proposed scheme trained by 70, 700, 1400, and 14000 DMRS sequences. 
Except for the Logistic Regression case in Fig.\ref{fig_11}, the fail probability decreases as the number of training sets increases. 
It is noted that it is easy to collect many DMRS sequences as the training sets in a certain short term, since the period of SSB transmission is normally 20ms and a UE can collect a DMRS sequence data per 20ms. 
(For instance, 30000 DMRS sequences can be collected in ten minutes which can be a sufficient time for training static UEs.)
The proposed schemes can consequently be further improved if it is trained by more DMRS sequences, and detect more accurately even at lower SNR. 
Thus, we can say that our proposed schemes can be adaptive to the current channel environment and become more powerful to detect the SSB index as data sets are accumulated. 
This characteristic differentiates the proposed schemes from the conventional scheme, which does not adapt to the channel and constantly performs worse in a severe channel environment.

\subsubsection{commercial gNB IOT Results}
In order to conduct the experiment in a commercial wireless environment, the channel-learning models are trained in prior by the DMRS sequences from the gNB emulation. Then test experiments are conducted with the commercial SK Telecom 5G cells and in two SNR environments. 
In the first environment, the SNR of the emulated gNB is –3.8dB and the SNR of the commercial cell is 6.42dB. 
This environment stands for training at lower SNR and testing at higher SNR.  
In the other environment, the SNR of the emulated gNB is 9.56dB and the commercial cell is 4.7dB. 
This stands for training at higher SNR and testing at lower SNR.

\begin{figure}
\includegraphics[width=\columnwidth]{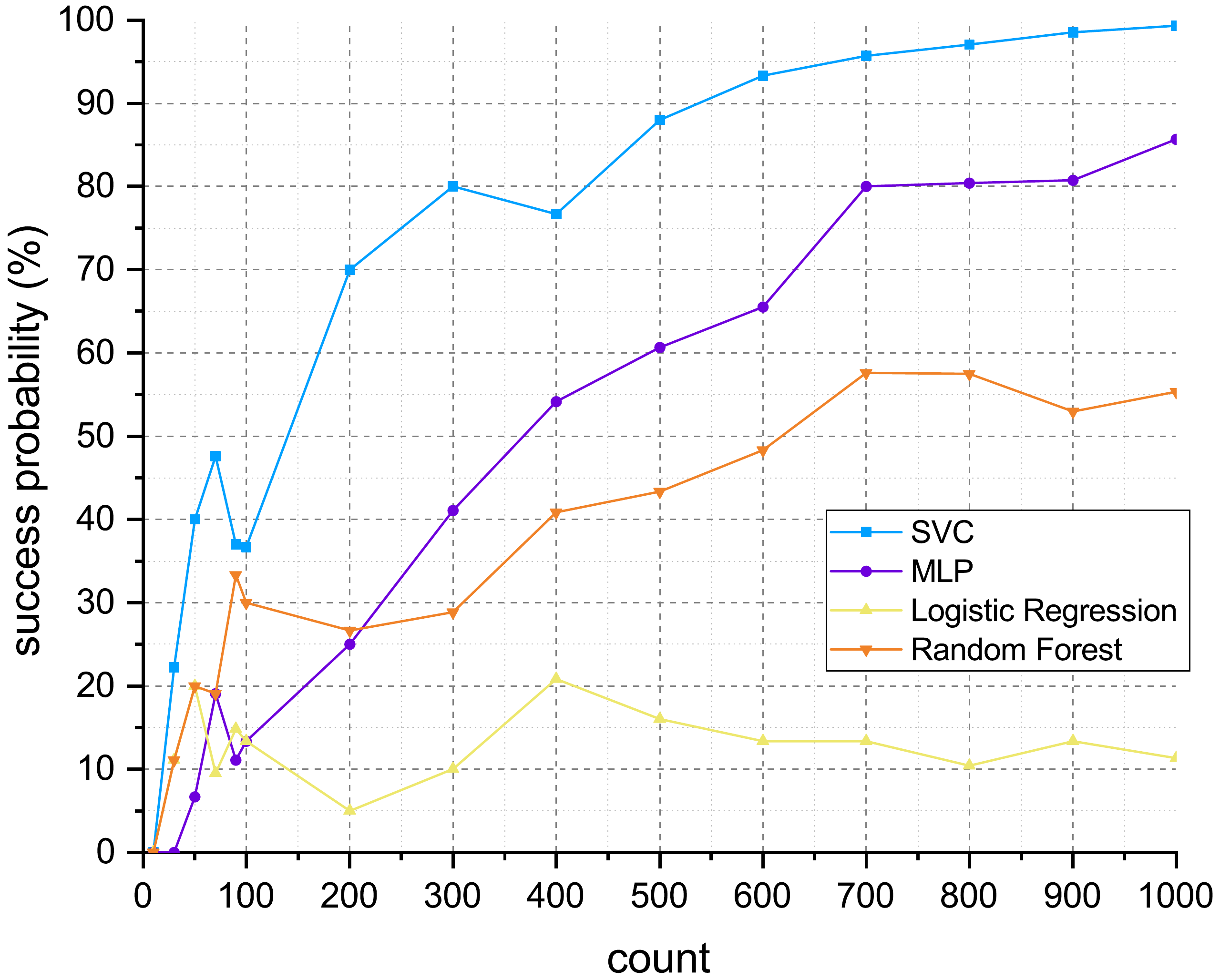}
\caption{The trend of the performance of the four channel-learning schemes over the time for the commercial 5G cell}
\label{fig_15}
\end{figure}

\begin{figure}
\includegraphics[width=\columnwidth]{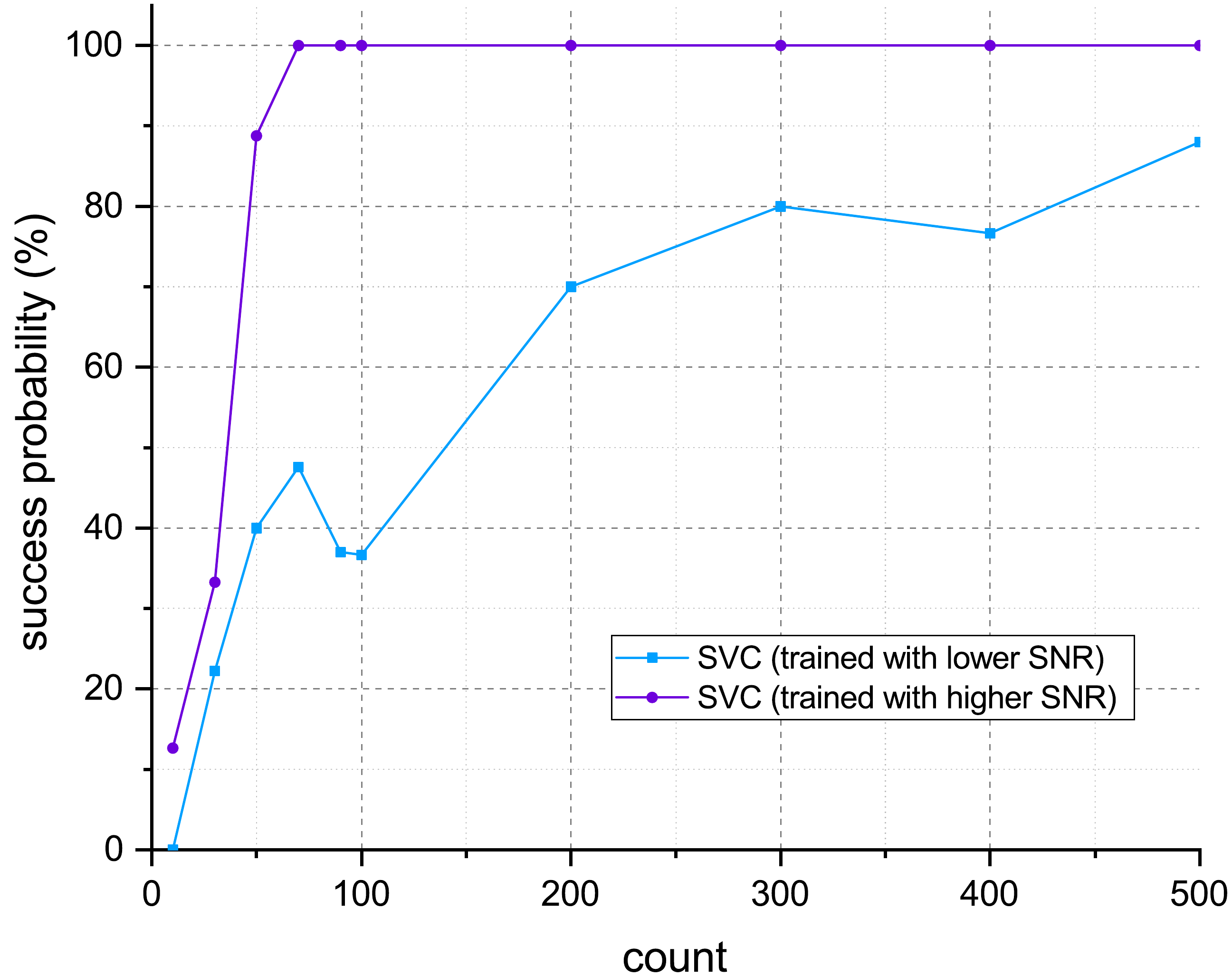}
\caption{The success probability of the SVC model in the two environments}
\label{fig_16}
\end{figure}

The results of the four channel-learning models in the first environment are shown in the Fig.\ref{fig_15} and the results of the SVC in both environments are shown in the Fig.\ref{fig_16}. 
The detection probability of the SVC converges to 100\% most quickly among the four channel-learning models as time goes by. 
In case of the SVC trained with 700 data sets, the detection probability in the first environment is 95\%, meanwhile the detection probabilities of the MLP classifier, Logistic Regression and Random Forest classifier are 80\%, 13\% and 57\%, respectively.

Fig.\ref{fig_16} also shows that if the SVC is trained at higher SNR, it takes less time for the convergence of the detection probability. 
As SNR gets higher, noise becomes less effective to the received signal, so the received DMRS sequence becomes less random. 
Therefore, the channel-learning model can quickly catch the characteristics of the input features and define the hypotheses. 
Also, once the performance of the channel-learning model is converged after training, the models keep performing well at any SNR. 
In the second environment, the SNR during testing is lower than in the first environment but the performance of the SVC is converged much faster. 
It means that the SNR during training has greater effect than the one during testing. 
In addition, the results reveal that even the training environment is different with the test environment, the channel-learning models can properly be trained for precise blind detection.
In a training overhead aspect, the results also imply that the channel-learning models do not need to be fully trained again once they have well trained under a certain environment.
The proposed schemes are therefore estimated to do the blind detection in a realistic channel environment.

\section{Conclusion}
In this paper, we proposed new blind detection schemes which find 5G beam indices by using channel-learning models. 
We chose the five channel-learning models which are applicable to the 5G cell search scenarios and designed the detailed procedures of the blind detection based on those models.
The performance of the proposed schemes was evaluated from the experiments in practical environments.
We collected 5G sampling data for training the channel-learning models by using the SDR testbed and the commercial 5G cells. 

The contribution of this research work can be summarized in the following two aspects: the application of ML models to 5G cell and beam index search scenarios and the performance evaluation of the channel-learning models in real channel environments. 
In the aspect of applying ML models, we can list up the following contributions in detail.
\begin{itemize}
	\item We defined and formulated the problem of the 5G beam index detection for which ML models are applicable.
	\item We designed pre-processing for input data of the channel-learning models. 
	\item We selected the five ML models and designed model architecture and parameters for each ML model. 
\end{itemize}

In the aspect of the performance evaluation, we can list up the following contributions in detail.
\begin{itemize}
	\item We built a practical testbed based on an SDR platform and provided a verification framework which enables to see the performance of 5G cell and beam index search in real wireless channel environments.
	\item We provided the performance of the blind beam index detection for the five ML models. We found that the SVC, MLP classifier and possibly the Ensemble Voting classifier perform better than the base-line scheme based on correlation, and this is more remarkable in a lower SNR environment.
	\item We provided the performance of the blind beam index detection for the number of training. We found that the SVC, which is the best performer, out-performs to the base-line scheme if it is trained with thousands of data set and can be further improved by training with more data sets. This implies that the channel-learning models can perform better than we deduced through the experiments if they are consistently trained.
	\item We showed that the channel-learning model performs fairly well even when the channel environment suffers a certain level of change. Even when the SNR changes, the channel-learning models can be sufficiently trained with hundreds of data sets and derive satisfactory detection results.
\end{itemize}

We can conclude that it is feasible to improve the performance of the 5G beam index detection by learning the channel with ML models.

\begin{IEEEbiography}[{\includegraphics[width=1in,height=1.25in,clip,keepaspectratio]{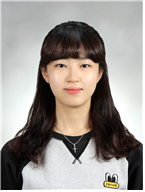}}]{Ji Yoon Han} 
(hanjiyoon17@sookmyung.ac.kr) is currently an undergraduate student at Sookmyung Women's University, Seoul, Korea. 
Since 2019 May, she is an undergraduate researcher at Bit Processing Laboratory (prof. Juyeop Kim).
She received best paper awards in 2020 winter and 2019 fall conferences of the Korean Institute of Communications and Information Sciences (KICS). 
Currently, her research interests includes Software-Defined Systems, Machine-Learning based communications systems and Internet of Things.
\end{IEEEbiography}

\begin{IEEEbiography}[{\includegraphics[width=1in,height=1.25in,clip,keepaspectratio]{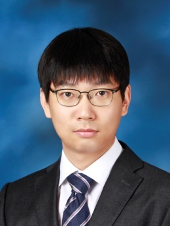}}]{Ohyun Jo} (ohyunjo@chungbuk.ac.kr) is currently an assistant professor of the Department of Computer Science at Chungbuk National University. He received his B.S., M.S., and Ph.D. degrees in electrical engineering from Korea Advanced Institute of Science and Technology (KAIST) in 2005, 2007, and 2011, respectively. From 2011 April to 2016 February, he was with Samsung Electronics in charge of research and development for future wireless communication systems, applications, and services. From 2016 March to 2017 July, he was a senior researcher at Electronics and Telecommunications Research Institute (ETRI) and from 2017 Aug. to 2018 Feb he was an assistant professor in the Department of Electrical Engineering. He has authored and co-authored more than 20 papers, and holds more than 140 registered and filed patents. During his appointment at Samsung, Dr. Jo is the recipient of numerous recognitions including Gold Prize in Samsung Annual Award, the Most Creative Researcher of the Year Award, the Best Mentoring Award, Major R\& D Achievement Award, and the Best Improvement of Organization Culture Award. His research interests include millimeter wave communications, next generation WLAN/WPAN systems, 5G mobile communication systems, military communications, Internet of Things, future wireless solutions/applications/services, and embedded communications ASIC design.
\end{IEEEbiography}

\begin{IEEEbiography}[{\includegraphics[width=1in,height=1.25in,clip,keepaspectratio]{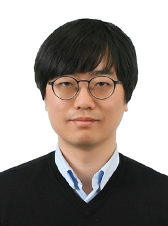}}]{Juyeop Kim} (jykim@sookmyung.ac.kr) is an assistant professor in the Department of Electronics Engineering, Sookmyung Women's University, Seoul, Korea. He received his B.S. and Ph.D. in electrical engineering from Korea Advanced Institute of Science and Technology (KAIST) in 2004 and 2010, respectively. From 2010 to 2011, he was with KAIST Institute IT Convergence Research Center in charge of research for 5G cellular system. From 2011 to 2013, he was with Samsung Electronics in charge of development and commercialization for 2G/3G/4G multi-mode mobile modem solution. From 2014 to 2018, he was with Korea Railroad Research Institute in charge of research and development for LTE-Railway(LTE-R), Public Safety LTE (PS-LTE) and railway IoT solutions. His current research interests is the applied wireless communications including mission critical communications, internet of things and software-defined modems.
\end{IEEEbiography}

\EOD

\end{document}